\documentclass[10pt,twocolumn,letterpaper]{article}

\usepackage{cvpr}
\usepackage{times}
\usepackage{epsfig}
\usepackage{graphicx}
\usepackage{amsmath}
\usepackage{amssymb}

\usepackage{authblk}

\usepackage[breaklinks=true,bookmarks=false]{hyperref}

\cvprfinalcopy 

\def\cvprPaperID{2637} 

\ifcvprfinal\fi
\begin{document}

\title{Cascade EF-GAN: Progressive Facial Expression Editing with Local Focuses}

\author[1]{Rongliang Wu}
\author[1]{Gongjie Zhang}
\author[1]{Shijian Lu\thanks{Corresponding author. This work is supported by Data Science \& Artificial Intelligence Research Centre, NTU Singapore.}}
\author[2]{Tao Chen}
\affil[1]{Nanyang Technological University}
\affil[2]{Fudan University}
\affil[ ]{\tt\small ronglian001@e.ntu.edu.sg, \{gongjiezhang,shijian.lu\}@ntu.edu.sg, eetchen@fudan.edu.cn}





\maketitle
\thispagestyle{empty}

\begin{abstract}
    Recent advances in Generative Adversarial Nets (GANs) have shown remarkable improvements for facial expression editing. However, current methods are still prone to generate artifacts and blurs around expression-intensive regions, and often introduce undesired overlapping artifacts while handling large-gap expression transformations such as transformation from furious to laughing. To address these limitations, we propose Cascade Expression Focal GAN (Cascade EF-GAN), a novel network that performs progressive facial expression editing with local expression focuses. The introduction of the local focuses enables the Cascade EF-GAN to better preserve identity-related features and details around eyes, noses and mouths, which further helps reduce artifacts and blurs within the generated facial images. 
    In addition, an innovative cascade transformation strategy is designed by dividing a large facial expression transformation into multiple small ones in cascade, which helps suppress overlapping artifacts and produce more realistic editing while dealing with large-gap expression transformations. Extensive experiments over two publicly available facial expression datasets show that our proposed Cascade EF-GAN achieves superior performance for facial expression editing.
\end{abstract}

\begin{figure}[!ht]
\begin{center}
\includegraphics[width=1.\linewidth]{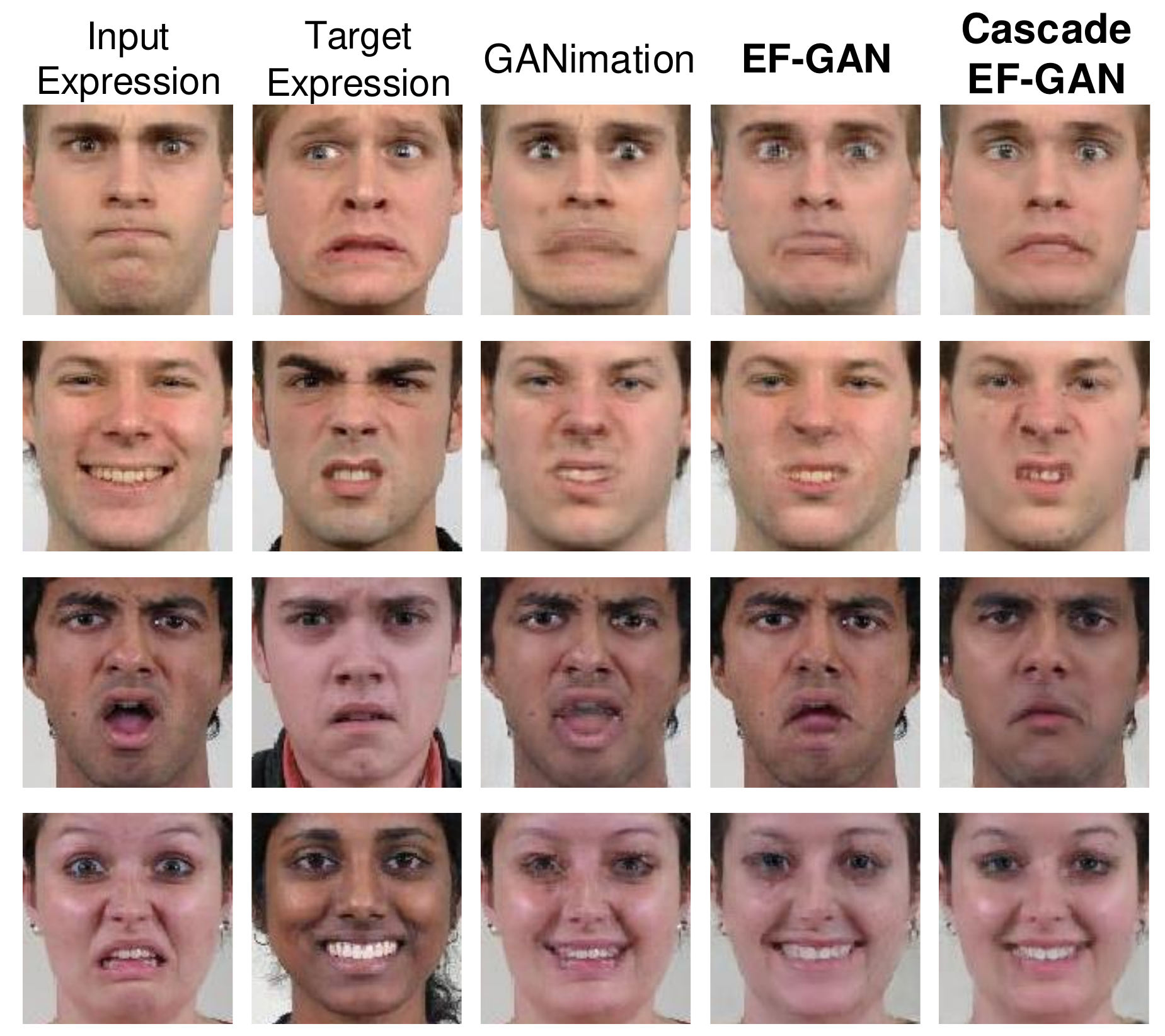}
\end{center}
\caption{
    Illustration of the proposed facial expression editing method: The introduction of local focuses (in \textbf{EF-GAN}) helps better preserve details and reduce blurs and artifacts. The proposed progressive facial expression editing (in \textbf{Cascade EF-GAN}) further helps remove overlapping artifacts and generates more realistic expression images.
}
\label{fig:fig1}
\end{figure}

\section{Introduction}
Facial expression opens a window to people's internal emotions and conveys subtle intentions~\cite{li2018deep} and there exists many research works on automatic facial expression recognition~\cite{zeng2018facial,wang2020suppressing,zhang2018joint,zhao2016peak,zhang2018joint}.  
In this day and age of digital media, facial expression editing, which transforms the expression of a given facial image to a target expression without losing identity properties, can potentially be applied in different areas such as photography technologies, movie industry, entertainment, etc.
It has been attracting increasing attention from both academia and industry.

Inspired by the recent success of Generative Adversarial Nets (GANs)~\cite{goodfellow2014generative}, several research works~\cite{qiao2018geometry,song2018geometry,ding2018exprgan,pumarola2018ganimation,choi2018stargan} have been reported and achieved very impressive facial expression editing results. On the other hand, existing methods are still facing a number of constraints. 
First, existing methods tend to generate incoherent artifacts and/or blurs, especially around those expression-rich regions such as eyes, noses and mouths. Second, existing methods tend to produce overlapping artifacts when the source facial expression has a large gap with the target facial expression, such as transformation from furious to laughing\footnote{Overlapping artifacts refer to the artifacts that original and target expressions are blended in the outputs.}.

The task of facial expression editing needs to maintain person identity. As humans, a natural way to identify facial images is to pay special attention to eyes, noses and mouths, largely because these regions contain rich identity-related information~\cite{althoff1999eye,hsiao2008two}. On the other hand, almost all GANs-based facial expression editing methods ~\cite{qiao2018geometry,song2018geometry,ding2018exprgan,pumarola2018ganimation,choi2018stargan} simply process the input facial image as a whole without paying special attention to local identity-related features, which could be one major reason why most existing methods generate incoherent artifacts and blurs around eyes, noses and mouths.

In addition, to the best of our knowledge, all existing GANs-based facial expression editing methods~\cite{qiao2018geometry,song2018geometry,ding2018exprgan,pumarola2018ganimation,choi2018stargan} perform a single-step transformation to the target expression. On the other hand, the single-step transformation often produces overlapping artifacts (around regions with large facial expression changes) while dealing with large-gap transformations due to the limitation of network capacity. 
Since facial expression changes are continuous by nature, a large-gap transformation should be better accomplished if the network decomposes it into a number of small transformations.

In this paper, we propose a novel {\it Cascade Expression Focal GAN (Cascade EF-GAN)} for progressive facial expression editing with local focuses. The Cascade EF-GAN consists of several identical EF-GAN modules in cascade that perform facial expression editing in a progressive manner. Specifically, an innovative cascade transformation strategy is designed which divides a large facial expression transformation into multiple small ones and performs facial expression transformation step-by-step progressively. Such progressive facial expression transformation helps suppress overlapping artifacts and achieve more robust and realistic expression editing while dealing with large-gap facial expression transformations. In addition, each EF-GAN module incorporates a number of pre-defined local focuses that capture identity-related features around eyes, noses and mouths, respectively. With the detailed identity-related features, the EF-GAN is capable of generating coherent facial expression images with much less artifacts. The results of our proposed Cascade EF-GAN are illustrated in Fig.~\ref{fig:fig1}.

The contributions of this work are threefold. First, we identify the importance of local focuses in facial expression editing, and propose a novel EF-GAN that captures identity-related features with several local focuses and mitigates the editing artifacts and blurs effectively. Second, we propose an innovative cascade design for progressive facial expression editing. The cascade design is robust and effective in suppressing overlapping artifacts while dealing with large-gap expression transformations. Third, extensive experiments show that the Cascade EF-GAN achieves superior facial expression editing quantitatively and qualitatively.

\begin{figure*}[t]
\begin{center}
\includegraphics[width=0.9\linewidth]{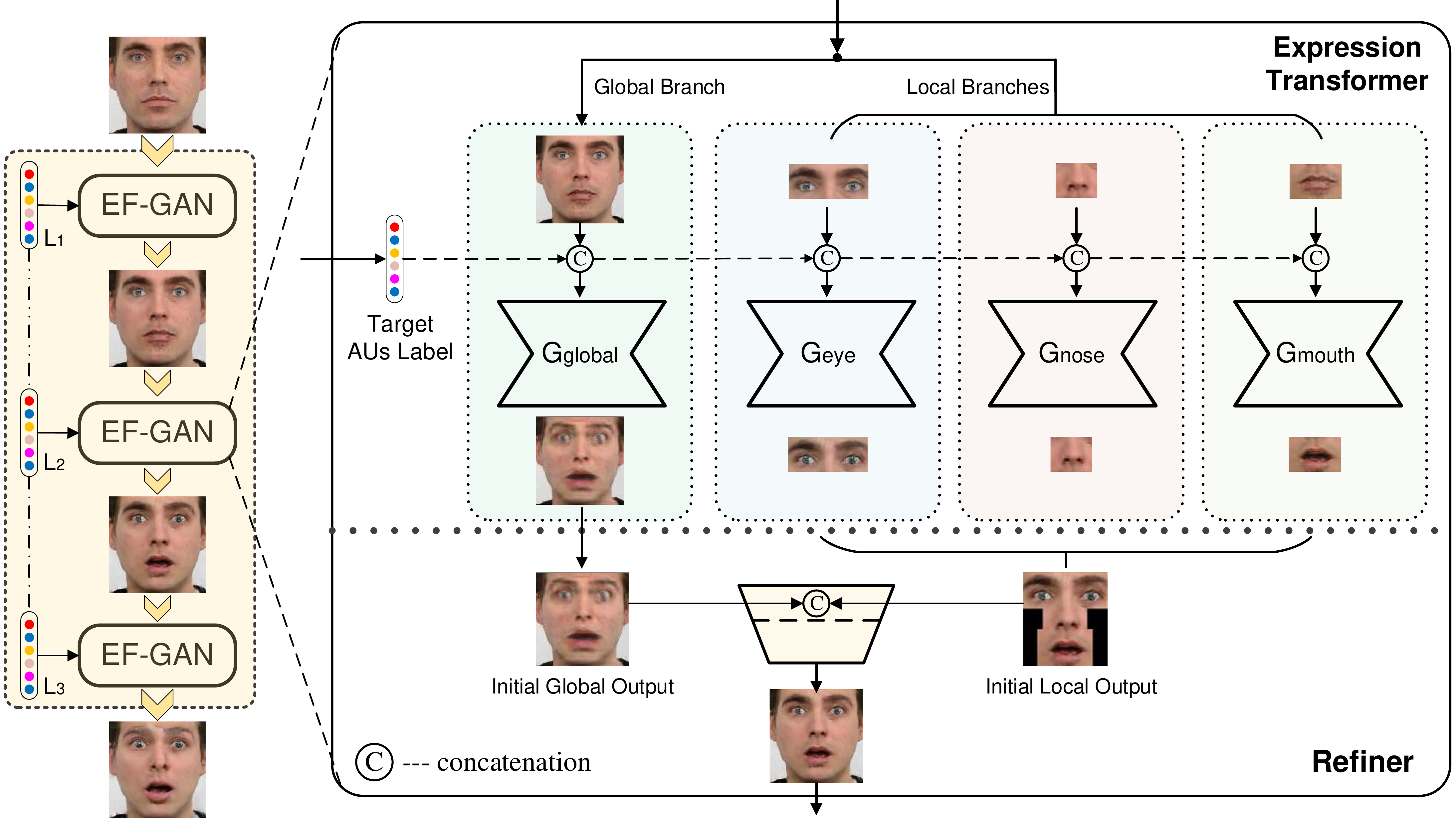}
\end{center}
\caption{
    Illustration of our Cascade EF-GAN: The workflow is shown on the left, and the details of each EF-GAN are shown in the zoom-in area. The expression editing is decomposed into 3 steps and handled by each EF-GAN progressively. The EF-GAN consists of an Expression Transformer and a Refiner: The former generates the initial editing of the whole facial image and three local facial regions, and the latter fuses the initial editing and refines it to generate the expression image as the final output.
}
\label{fig:overall_arc}
\end{figure*}

\section{Related Work}

{\bf Generative Adversarial Nets:} Generative Adversarial Nets (GANs) are powerful generative models that simultaneously train a generator to produce realistic fake images and a discriminator to distinguish between real and fake images. One active research topic is Conditional GANs~\cite{mirza2014conditional} that include conditional information to control the generated images. In addition, CycleGAN~\cite{zhu2017unpaired} adopts cycle-consistency loss and achieves image-to-image translation with well preserved key attributes. 
GANs have demonstrated their powerful capabilities in different computer vision tasks such as natural image synthesis~\cite{brock2018large,karras2017progressive}, 
image style translation~\cite{isola2017image,zhu2017unpaired,kim2017learning,liang2017generative,wang2018high}, 
super-resolution~\cite{ledig2017photo,wang2018cascaded,sajjadi2017enhancenet},
image inpainting~\cite{yeh2017semantic,yu2018generative,pathak2016context,yan2018shift}, 
facial attributes editing~\cite{choi2018stargan,zhang2018generative,xiao2018elegant,shen2017learning,lu2018attribute,chen2019homomorphic,chen2019semantic}, 
face image synthesis~\cite{lample2017fader,huang2017beyond,yin2017towards,tran2017disentangled,zhao20183d}, etc. 
The GAN-generated images have also been applied to different computer vision tasks~\cite{zhao20183d,huang2017beyond,zhao2017dual,li2017generative}. 
Our Cascade EF-GAN is designed to perform facial expression editing, with conditional variables to control the target expressions and cycle-consistency to preserve identity information.

{\bf Facial Expression Editing:} Facial expression editing is challenging as it requires high-level understanding of input facial images and prior knowledge about human expressions. 
Compared with general facial attributes editing which only considers appearance modification of specific facial regions~\cite{zhang2018generative,li2016deep,shen2017learning}, facial expression editing is a more challenging task as it often involves large geometrical changes and requires to modify multiple facial components simultaneously.
Very impressive progress has been achieved with the prevalence of GANs in recent years. For example, G2-GAN~\cite{song2018geometry} and GCGAN~\cite{qiao2018geometry} adopt facial landmarks as geometrical priors to control the intensity of the generated facial expressions, where ground-truth images are essential for extracting the geometrical information. 
ExprGAN~\cite{ding2018exprgan} introduces an expression controller to control the intensity of generated expressions, but it requires a pre-trained face recognizer for preserving the identity information. StarGAN~\cite{choi2018stargan} can translate images across domains with a single model and preserve identity features by minimizing a cycle loss, but it can only generate discrete expressions. GANimation~\cite{pumarola2018ganimation} adopts Action Units as expression labels and can generate expressions in continuous domain. 
It also incorporates attention to better preserve the identity information. 
However, it tends to generate artifacts and blurs and cannot handle large-gap expression transformations well. 

Instead of generating expressions on a whole face image as existing GAN-based methods, our proposed Cascade EF-GAN includes local focuses on eyes, nose and mouth regions that help suppress artifacts and preserve details clearly. In addition, the cascade strategy edits expressions in a progressive manner, which is able to suppress the overlapping artifacts effectively while dealing with transformations across very different expressions.

\section{Proposed Methods}

Fig.~\ref{fig:overall_arc} shows the overall framework of our proposed Cascade EF-GAN. As Fig.~\ref{fig:overall_arc} shows, the Cascade EF-GAN consists of multiple EF-GANs in cascade that performs expression editing in a progressive manner. Each EF-GAN shares the same architecture, which consists of an Expression Transformer and a Refiner. Specifically, several pre-defined local focuses branches are incorporated into each EF-GAN module for better preserving identity-related features and details around eyes, noses and mouths. More details are to be discussed in the ensuing subsections.

\subsection{EF-GAN with Attention-Driven Local Focuses}

The generative model within EF-GAN consists of an Expression Transformer that performs expression editing with local focuses, and a Refiner that fuses the outputs from the Expression Transformer and refines the final editing.

\noindent {\bf Expression Transformer.} Fig.~\ref{fig:overall_arc} shows the architecture of our Expression Transformer. Different from existing methods~\cite{qiao2018geometry,song2018geometry,ding2018exprgan,pumarola2018ganimation,choi2018stargan} that employ a single global branch to process the facial image, our Expression Transformer incorporates three additional local branches with pre-defined focuses on local regions around eyes, noses and mouths, respectively. The motivation is that convolutional kernels are shared across all spatial locations, but each facial region has distinct identity-related features. Simply processing a facial image as a whole with one set of convolutional kernels is thus not sufficient to capture identity-related details around each facial region. Inspired by~\cite{huang2017beyond,zhao20183d}, our Expression Transformer tackles this challenge by processing facial images in both global and local branches, where the global branch captures global facial structures and local branches focus on more detailed facial features.

Specifically, the Expression Transformer takes a facial image and a target expression label as input. Similar to GANimation~\cite{pumarola2018ganimation}, we adopt the Facial Action Coding System (FACS)~\cite{friesen1978facial} that encodes expressions to Action Units (AUs) which can be extracted by using the open-source OpenFace~\cite{baltrusaitis2018openface}. We adopt the continuous AUs intensity as AUs labels to supervise the editing process. Given a source facial expression image, local focuses are first applied to the eyes, nose and mouth regions by cropping the corresponding local image patches. The landmarks for each local focus are also acquired by OpenFace~\cite{baltrusaitis2018openface}. The global facial image and its local patches are then fed to the corresponding branches of the Expression Transformer for expression editing. Note all branches share similar network architectures without sharing weights.

\begin{figure}[t]
\begin{center}
\includegraphics[width=0.84\linewidth]{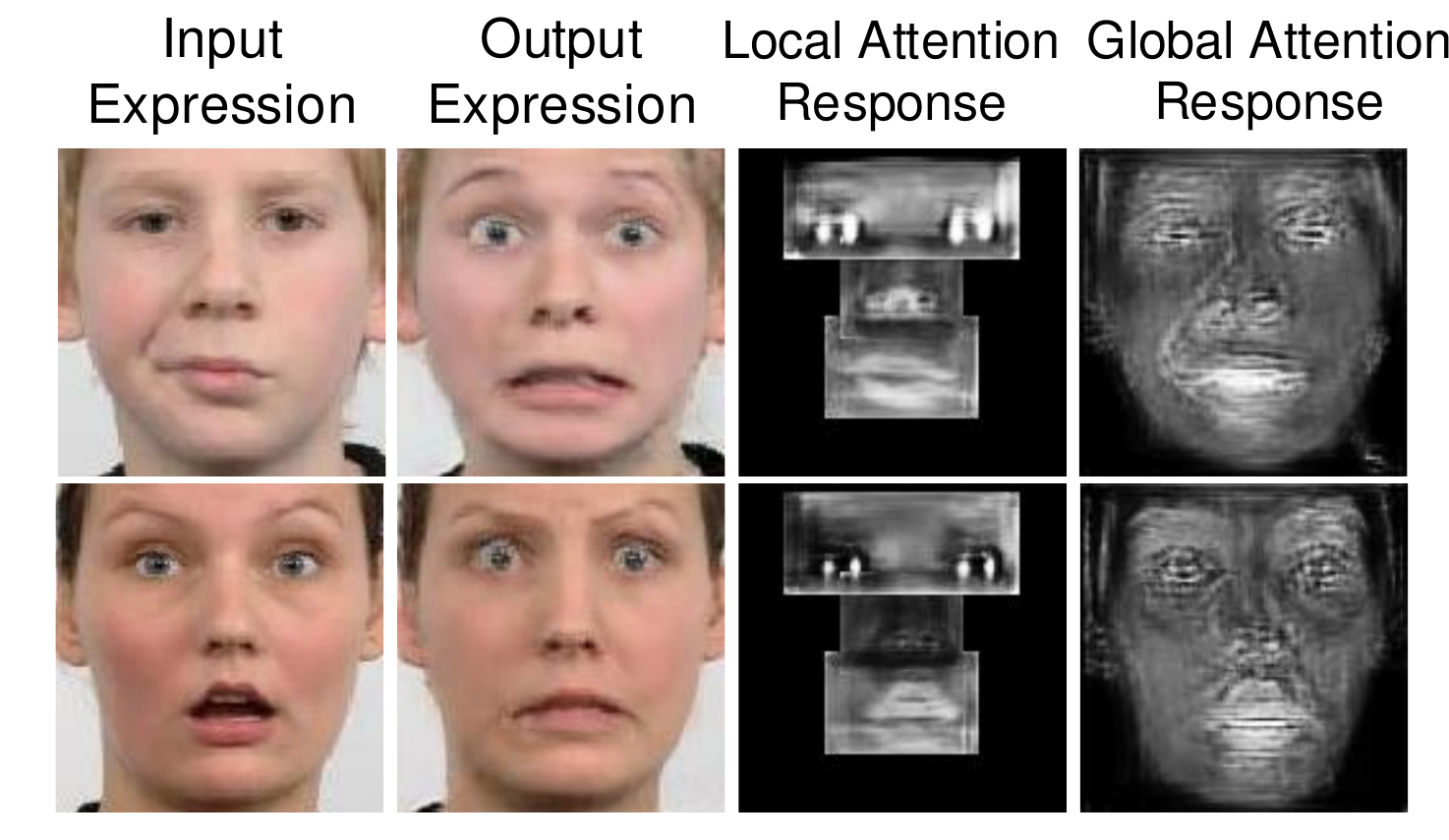}
\end{center}
\caption{
     Illustration of local and global attention: global attention tends to capture the most salient expression changes, e.g. the mouth region; local attention captures fine expression changes at local regions, e.g. the eyes region. Note local attention responses are stitched for presentation purpose.
}
\label{fig:attention}
\end{figure}

We also incorporate attention in global and local branches for better details capture and artifacts suppression. 
The use of visual attention has been investigated in GANimation~\cite{pumarola2018ganimation}, where attention was designed to guide the network to focus on transforming expression-related regions. On the other hand, applying attention in a single global image often introduces vague attention responses as illustrated in column 4 of Fig.~\ref{fig:attention}. This is because the global attention tends to focus on the most salient changes, e.g. the mouth regions in Fig.~\ref{fig:attention}, whereas fine changes around the eyes and nose are not well attended. The exclusive attention in the aforementioned local branches helps to achieve sharper responses at local regions as shown in column 3.

Specifically, each branch outputs color feature maps $M_C$ and attention map $M_A$. With the original input image $I_{in}$, the initial output of each branch is generated by:
\begin{equation}
    \mathcal{I}_{init} = {{M}_{A} \otimes {M}_{C} + (1 - {M}_{A}) \otimes {I}_{in}}, 
    \label{formula:attention}
\end{equation}
where $\otimes$ denotes element-wise multiplication.
This strategy eases the learning process greatly as the network does not need to output the initial results directly under the constraint of facial images statistics.

The Expression Transformer just generates initial expression editing as shown in Fig.~\ref{fig:overall_arc}. Specifically, the global branch captures global facial structures and features but generates blurs and artifacts around local regions due to the miss of local details. The local branches preserve local details better but they miss the big picture of the whole facial expression. The outputs of the two types of branches are therefore sent to the Refiner for fusion and further improvement. 

\noindent {\bf Refiner.} The Refiner is responsible for fusing the outputs of different branches of the Expression Transformer and generating the final expression editing. As Fig.~\ref{fig:overall_arc} shows, the outputs of the three local branches are first stitched into a single image according to their respective locations within a facial image. The stitched image is then concatenated with the output of the global branch and fed to the Refiner to generate the final expression editing.


\begin{figure}[t]
\begin{center}
\includegraphics[width=1\linewidth]{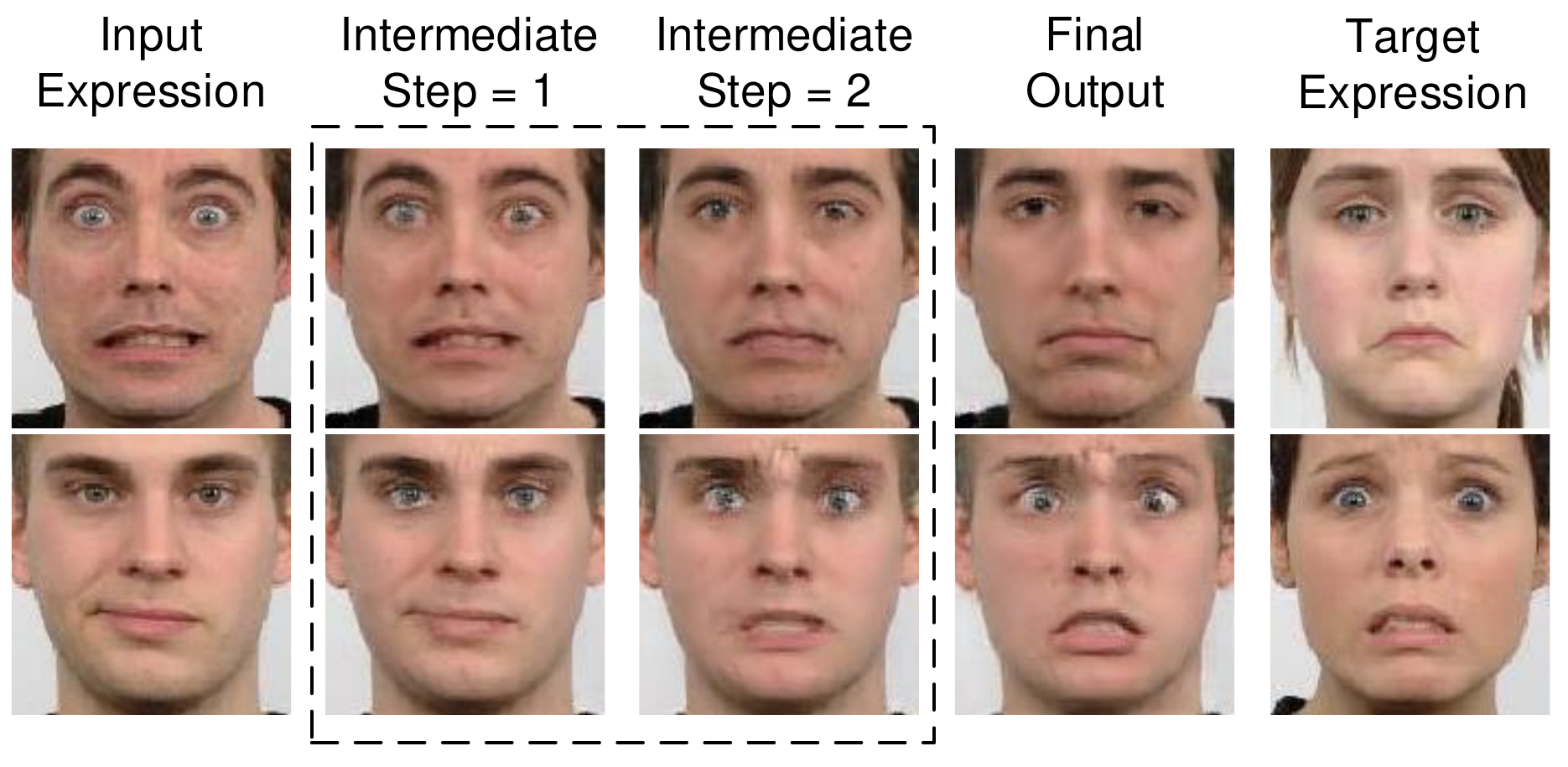}
\end{center}
\caption{
     Illustration of intermediate and final editing by our Cascade EF-GAN: the progressive editing helps suppress overlapping artifacts and produce more realistic editing while dealing with large-gap expression changes.
}
\label{fig:cascade_examples}
\end{figure}

\subsection{Cascade Facial Expression Transformation}

\noindent {\bf Cascade Framework.} Given an input facial image, the aforementioned EF-GAN is able to generate high-fidelity expression editing in most cases. On the other hand, our study shows that EF-GAN tends to produce overlapping artifacts around the regions with large expression changes while dealing with large-gap expression transformations.
We refer large-gap expression transformations to those transformations that involve great appearance and geometrical modifications for editing the expression, such as transformation from furious to laughing.
To address this constraint, we propose Cascade EF-GAN that performs expression editing in a progressive manner. Specifically, the Cascade EF-GAN decomposes a large-gap expression transformation into multiple small ones and performs large-gap expression transformations in cascade. It allows better preservation of facial structures and identity-related features as well as robust handling of large-gap facial transformations. 

As Fig.~\ref{fig:overall_arc} shows, the cascade expression editing is achieved by cascading multiple EF-GANs together, where the expression image from the previous EF-GAN is fed to the ensuing one as input for further editing. We empirically use 3 EF-GANs and Fig.~\ref{fig:cascade_examples} shows intermediate and final expression editing by the proposed Cascade EF-GAN. As Fig.~\ref{fig:cascade_examples} shows, the challenging large-gap expression editing is accomplished progressively in multiple steps, leading to realistic facial images of target expressions smoothly. 


\noindent {\bf Intermediate Supervision:} Another issue in implementing the progressive editing is how to include supervision information into each intermediate step. With the AU labels of input expression and the target expression, the straightforward approach is to generate intermediate AUs by linear interpolation. However, such interpolated AUs may not reside on the manifold of natural AUs and lead to weird synthesis. We address this issue by training an Interpolator to produce the intermediate AUs. Specifically, we first generate pseudo intermediate targets by linear interpolation and extract the residuals between the pseudo targets and the original AUs labels of input expression. The original AUs labels and residuals are then fed to the Interpolator to produce the intermediate AUs for providing supervision for the intermediate expression. 
Besides, a discriminator is trained to maximize the Wasserstein distance between the real and interpolated AUs while the Interpolator is trained to minimize the distance between them, so that the interpolated ones cannot be distinguished from real samples. Note all EF-GANs use the same AUs Interpolator. 

\subsection{Learning the Model}

\noindent {\textbf{Loss Function}}
The loss function for training the proposed EF-GAN contains five terms:
1) the adversarial loss for improving the photo-realism of the synthesized facial expression images to make them indistinguishable from real samples;
2) the conditional expression loss to ensure generated facial expression images to align with the provided target AUs labels;
3) the content loss for preserving the identity information and consistency of the image content.
4) the attention loss to encourage the attentive module to produce sparse attention map and pay attention to the regions that really need modification. 
5) the interpolation loss to constrain the interpolated AUs label has desired sematical meaning and resides on the manifold of natural AUs.
The overall objective function is expressed as:
\begin{equation}
    \begin{aligned}
        \mathcal{L} = \mathcal{L}_{adv}
                        +\lambda_{1} \mathcal{L}_{cond}
                        +\lambda_{2} \mathcal{L}_{cont}
                        +\lambda_{3} \mathcal{L}_{attn}
                        +\lambda_{4} \mathcal{L}_{interp}
        \label{formula:overall_loss}
    \end{aligned}
\end{equation}
where $\lambda_{1}$, $\lambda_{2}$, $\lambda_{3}$ and $\lambda_{4}$ are the hyper-parameters.
In Cascade EF-GAN, the total loss is the sum of the loss
of each EF-GAN with equal weights.
Due to the limit of the paper length, please refer to the supplementary materials for the detailed loss and network architecture.

\noindent {\bf Training Scheme:} It is difficult to obtain good expression editing if we directly cascade multiple EF-GAN modules and train them from scratch. We conjecture that this is largely due to the noisy facial images from the early-stage EF-GAN modules. Taking such noisy facial images as input, the later stages of the Cascade EF-GAN can be easily affected and produce degraded editing. In addition, the undesired editing will be accumulated, which makes network parameters difficult to optimize.

We design a simple yet effective scheme to address this issue. Specifically, we first train a single EF-GAN to perform a single-step facial expression transformation. Then, we use the weights of the well-trained EF-GAN to initialize all following EF-GAN in the cascade and fine-tune all network parameters end-to-end. With this training scheme, each EF-GAN module in the cascade will have good initialization, thus the intermediate facial expression images become useful to enable later stages to learn meaningful expression transformation information.

\section{Experiments}

\begin{figure*}[t]
\begin{center}
\includegraphics[width=1.\linewidth]{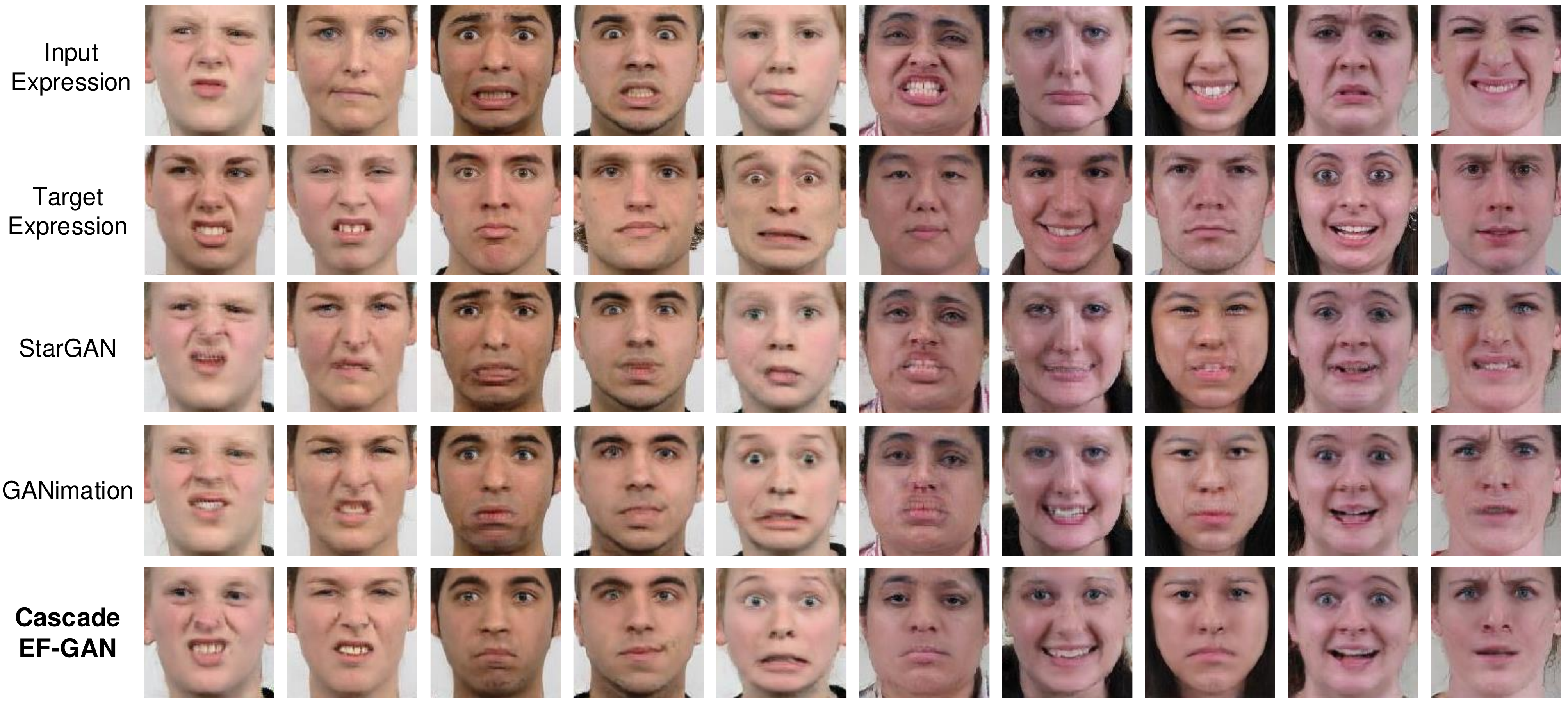}
\end{center}
\caption{
    Expression editing by Cascade EF-GAN and state-of-the-art methods: Columns 1-5 show the editing of RaFD images, and the rest shows that of CFEED. Our Cascade EF-GAN produces more realistic editing with better details and less artifacts.
}
\label{fig:exp_both}
\end{figure*}

\subsection{Datasets}
\label{Datasets}
The Cascade EF-GAN is evaluated over Radboud Faces Dataset (RaFD)~\cite{langner2010presentation} and Compound Facial Expressions of Emotions Dataset (CFEED)~\cite{du2014compound}.
RaFD consists of 8,040 expression images collected from different angles. We use the facial images captured by the $90^{\circ}$ camera, leading to 1,608 facial expression images.
CFEED contains 5,060 compound expression images collected from 230 participants. We randomly sample 90\% for training and the rest are used for testing.

In our experiments, we crop the images into 128 $\times$ 128 patches with faces in the center. The sizes of the three local patches (i.e. patches of eyes, nose and mouth) are fixed to 40 $\times$ 92, 40 $\times$ 48 and 40 $\times$ 60, respectively. The center of each patch is the average position of the corresponding key points over all training samples. 

\subsection{Qualitative Experimental Results}
The proposed Cascade EF-GAN is evaluated over two publicly available facial expression datasets described in previous section. Fig.~\ref{fig:exp_both} shows qualitative experimental results, where images in columns 1-5 are from the RaFD dataset and images in columns 6-10 are from the CFEED dataset. Each column includes an expression editing task, along with facial expression editing by state-of-the-art methods -- StarGAN~\cite{choi2018stargan} and GANimation~\cite{pumarola2018ganimation} as well as our proposed Cascade EF-GAN. 

As Fig.~\ref{fig:exp_both} shows, state-of-the-art methods are prone to generate blurs and artifacts and even corrupted facial expressions around eyes, nose and mouth regions. Our Cascade EF-GAN instead generates more realistic facial expressions with much less blurs and artifacts, and its generated images are also much clearer and sharper. The better synthesis is largely attributed to the inclusion of attention-driven local focuses that helps to better preserve identity-related features and details in the corresponding facial regions. In addition, state-of-the-art methods tend to produce overlapping artifacts while handling large-gap expression transformations. Our Cascade EF-GAN instead suppresses such overlapping artifacts effectively, largely due to our cascade design that performs human-like progressive expression transformation rather than a single-step one. More results are provided in the supplementary materials.

\begin{table}[t]
    \caption{
            Quantitative comparison with state-of-the-art on RaFD and CFEED datasets with facial expression recognition accuracy. 
            }
    \begin{center}
        \begin{tabular}{|c|c|c|c|c|}
          \hline
          Dataset & Method & R & G & R + G \\
          \hline
          {}    & StarGAN~\cite{choi2018stargan}    & {}    & 82.37      & 88.48 \\
          \cline{2-2} \cline{4-5}
          RaFD  & GANimation~\cite{pumarola2018ganimation} & 92.21 & 84.36      & 92.31 \\
          \cline{2-2} \cline{4-5}
          {}    & Ours       & {}    & \bf{89.38} & \bf{93.67} \\
          \hline
          {}    & StarGAN~\cite{choi2018stargan}    & {}    & 77.80      & 81.87 \\
          \cline{2-2} \cline{4-5}
          CFEED & GANimation~\cite{pumarola2018ganimation} & 88.23 & 79.46      & 84.42 \\
          \cline{2-2} \cline{4-5}
          {}    & Ours       & {}    & \bf{85.81} & \bf{89.25} \\
          \hline
        \end{tabular}
    \end{center}
    
    \label{table:cls_accuracy}
\end{table}

\begin{table}[t]
    \caption{
            Quantitative comparison with state-of-the-art on RaFD and CFEED datasets with PSNR (higher is better) and FID (lower is better).
            }

    \begin{center}
        \begin{tabular}{|c|c|c|c|c|c|}
          \hline
          {}             & \multicolumn{2}{|c|}{RaFD} & \multicolumn{2}{|c|}{CFEED} \\
          \hline
          {}                                         & PSNR$\uparrow$          & FID$\downarrow$            & PSNR$\uparrow$           & FID$\downarrow$   \\
          \hline
          StarGAN~\cite{choi2018stargan}             & 19.82      & 62.51       & 20.11       & 42.39 \\
          \cline{1-5}
          GANimation~\cite{pumarola2018ganimation}   & 22.06      & 45.55       & 20.43       & 29.07  \\
          \cline{1-5}
          Cascade EF-GAN                             & \bf{23.07} & \bf{42.36}  & \bf{21.34}  & \bf{27.15}\\
          \hline
        \end{tabular}
    \end{center}
    
    \label{table:PSNR_FID}
\end{table}

\subsection{Quantitative Experimental Results}
\label{sec:quantitative_exp}

\noindent \textbf{Expression Classification Accuracy:} 
We follow the evaluation method of StarGAN~\cite{choi2018stargan} and ExprGAN~\cite{ding2018exprgan} for quantitative evaluations. Specifically, we first train different expression editing models on the training set and perform expression editing on the same, unseen testing set. Then the generated images are evaluated in different expression recognition tasks. A higher classification accuracy indicates more accurate and realistic expression editing. 

Two classification tasks are designed to evaluate the quality of the generated images: 1) train an expression classifier by using the original training images and apply the classifier to classify the expression images that are generated by different editing methods; 2) train classifiers by combining the natural and generated expression images to classify the original test set images. The first task evaluates whether the generated images lie in the manifold of natural expressions, and the second evaluates whether the generated images help train better classifiers.

\begin{figure*}[t]
\begin{center}
\includegraphics[width=1.\linewidth]{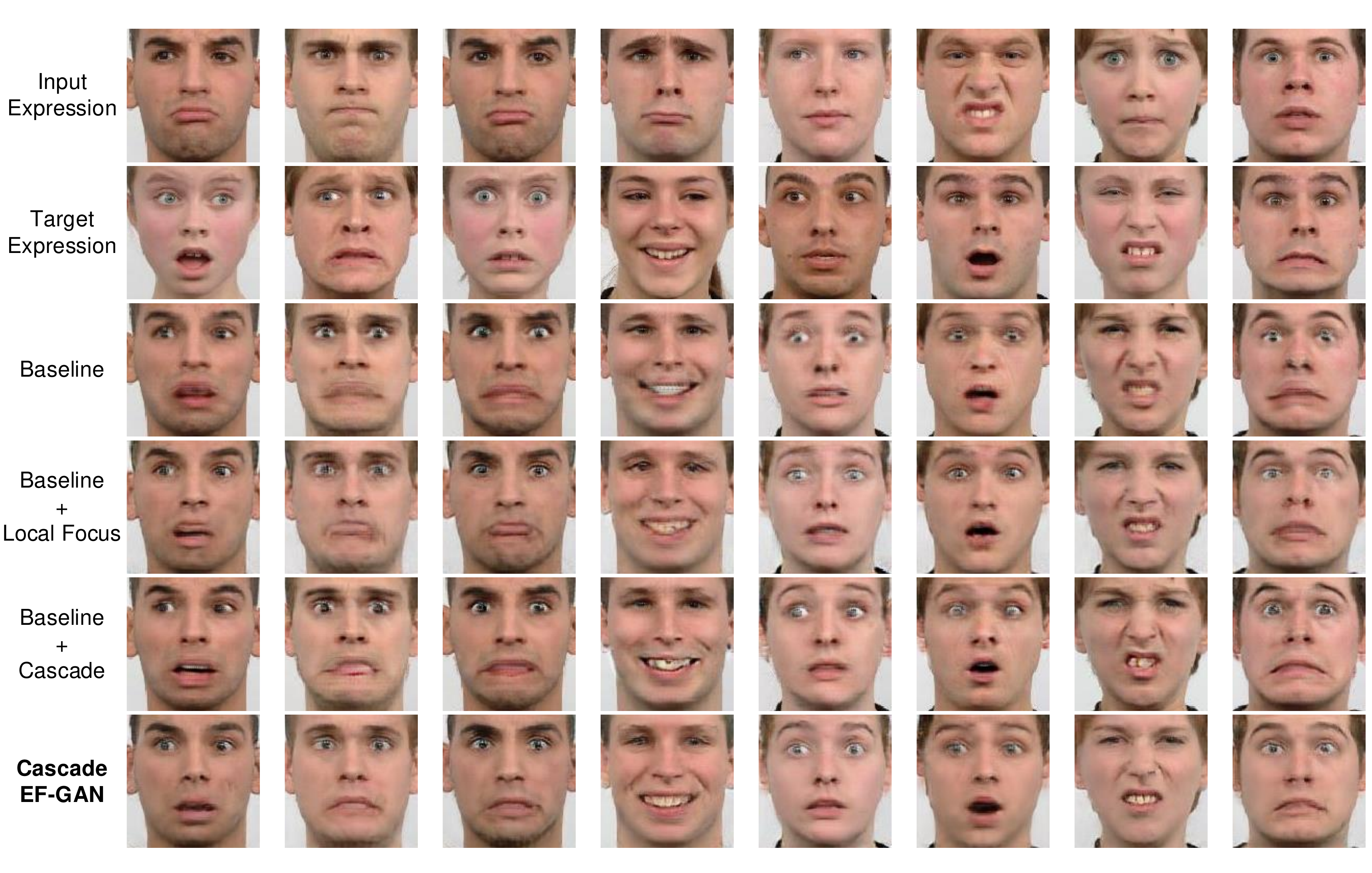}
\end{center}
\caption{
    Ablation study of our Cascade EF-GAN over RaFD images: The inclusion of local focuses and cascade strategy helps to produce sharper expression images with better preserved details and less artifacts. Zoom-in may be needed for more details. 
}
\label{fig:exp_ablation}
\end{figure*}

Table~\ref{table:cls_accuracy} shows the expression classification accuracy on RaFD and CFEED (only seven primary expressions are evaluated for CFEED). Specifically, \textbf{R} means to train a classifier with original training set images then apply it to recognize the expression of testing set images. \textbf{G} means to use the same classifier (as in \textbf{R}) to recognize the expression of the generated images. \textbf{R + G} means to train classifiers by combining real and the generated images of different methods then apply them to recognize the expression of testing set images.
As the table shows, our cascade EF-GAN achieves the highest accuracy in the first task, with 89.38$\%$ on RaFD and 85.81$\%$ on CFEED, showing its superiority in generating more realistic expression images. Additionally, it can help to train more accurate expression classifiers, where the accuracy is improved by 1.46$\%$ and 1.02$\%$ on RaFD and CFEED when our generated images are combined with real images in classifier training. As a comparison, StarGAN~\cite{choi2018stargan} and GANimation~\cite{pumarola2018ganimation} generated images tend to degrade the classification, probably due to the artifacts and blurs within their generated images.

\begin{table}[t]
    \caption{
             Ablation study of classification accuracy on RaFD.
        }

    \begin{center}
        \begin{tabular}{|c|c|c|}
          \hline
           Method                & G     & R + G\\
          \hline
           Baseline              & 84.61        & 92.37    \\
          \hline
           Baseline + Local Focuses  & 86.59        &  92.50   \\           \hline
           Baseline + Cascade    & 87.94        &  92.77    \\           \hline
           Cascade EF-GAN       & \bf{89.38}   & \bf{93.67}  \\
          \hline
        \end{tabular}
    \end{center}
    
    \label{table:ablation_evaluation}
\end{table}

\noindent \textbf{PSNR and FID:} 
We also evaluate the quality of the generated images with peak signal-to-noise ratio (PSNR)~\cite{huynh2008scope} and Fr\'{e}chet Inception Distance (FID)~\cite{heusel2017gans} metrics. 
The PSNR is computed over synthesized expressions and corresponding expressions of the same identity while the FID scores are calculated between the final average pooling features of a pretrained inception model~\cite{szegedy2017inception} of real faces and the synthesized faces.
As shown in Table~\ref{table:PSNR_FID}, our proposed Cascsde EF-GAN outperforms the state-of-the-art method by 1.01/ 3.19 under the measurement of PSNR and FID on RaFD dataset, and 0.91/ 1.92 on CFEED, respectively.

\begin{figure*}[t]
\begin{center}
\includegraphics[width=1\linewidth]{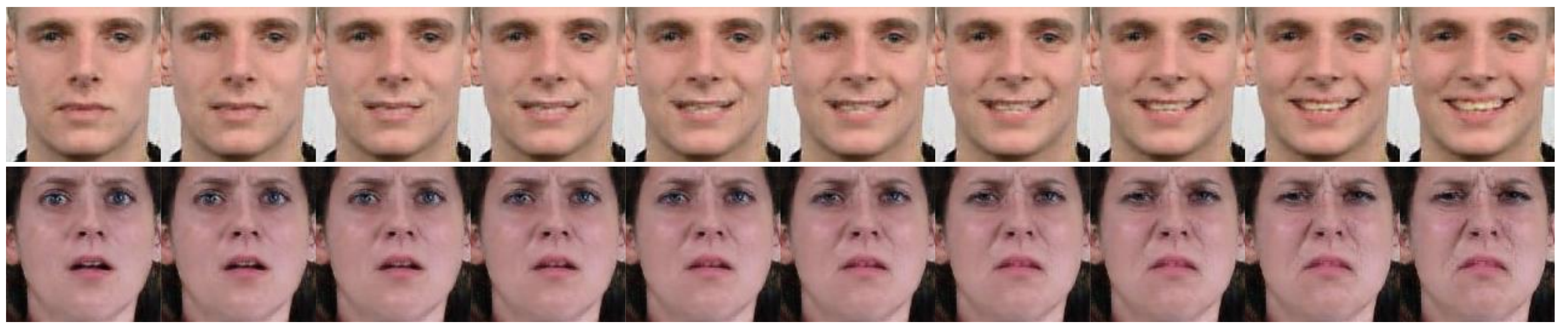}
\end{center}
\caption{
    Continuous expression editing by Cascade EF-GAN: For input images in Column 1, the rest gives continuous editing.
}
\label{fig:smooth_transistion}
\end{figure*}

\begin{figure*}[t]
\begin{center}
\includegraphics[width=1\linewidth]{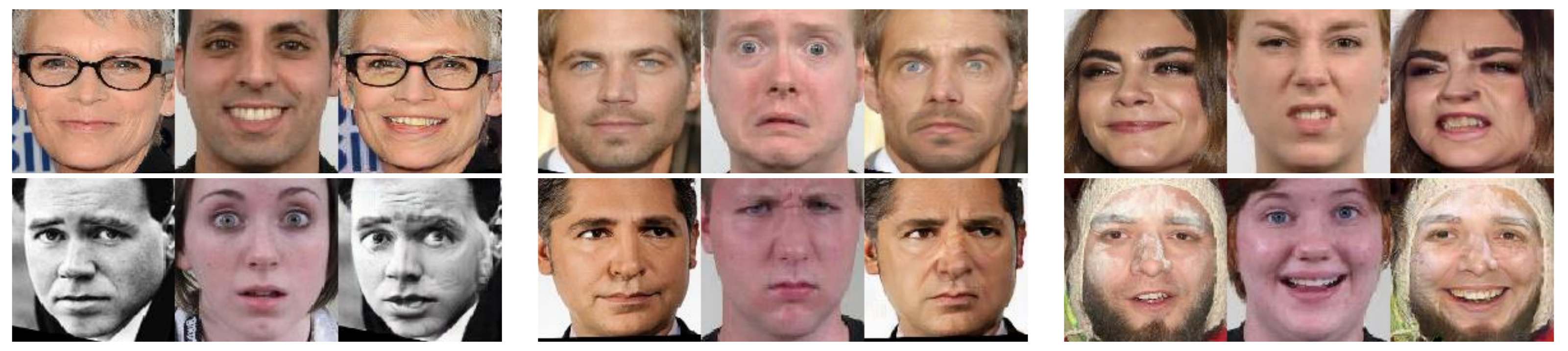}
\end{center}
\caption{
    Facial expression editing by the Cascade EF-GAN on wild images: In each triplet, the first column is input facial image, the second column is the image with desired expression and the last column is the synthesized result.
}
\label{fig:exp_wild}
\end{figure*}

\subsection{Ablation Study}
\label{sec:ablation_study}
We perform ablation studies over the RaFD dataset to study the contributions of our proposed local focuses and cascade designs. Several models are trained including: 1) \textbf{Baseline} where only global attention is adopted as GANimation~\cite{pumarola2018ganimation}; 2) \textbf{Baseline + Local Focuses} (i.e. EF-GAN) that includes the local focuses branches into the \textbf{Baseline}; 3) \textbf{Baseline + Cascade} that includes the progressive editing (with 3 EF-GAN modules) into the \textbf{Baseline}; and 4) \textbf{Cascade EF-GAN} that includes both progressive editing and local focuses as illustrated in Fig.~\ref{fig:overall_arc}.

Fig.~\ref{fig:exp_ablation} shows qualitative results. Each column represents an expression editing task and the corresponding editing by the aforementioned models. It is obvious that the {\bf Baseline} tends to lose details around eyes and mouths, resulting in blurs, artifacts and even corruptions there. The generated expressions are not well aligned with target expressions either for a number of sample images. The {\bf Baseline + Local Focuses} reduces artifacts and corruptions greatly, and generates clearer and sharper expression images. The inclusion of the cascade strategy in \textbf{Baseline + Cascade} helps better maintain the identity features and face structure, and the generated expressions also align better with the target expressions. This is largely because the cascade design mitigates the complexity of large-gap changes by decomposing them into smaller steps. Finally, the {\bf Cascade EF-GAN} which includes both the cascade design and local focuses is able to generate clean and sharp facial expressions that are well aligned with both target expressions and input identities, clearly better than all other models. This shows that the proposed local focuses and the cascade editing strategy are complimentary to each other.

We also conduct quantitative experiments to evaluate each proposed component in the Cascade EF-GAN. Table~\ref{table:ablation_evaluation} shows experimental results. The quantitative experimental results 
further verify the effectiveness of the proposed local focuses and progressive transformation strategy.

\subsection{Discussion}
\noindent \textbf{Continuous Expression Editing:} Our Cascade EF-GAN can be easily adapted to generate continuous expressions. Given the source and target AUs, intermediate AUs of different stages can be derived with the Interpolator. Continuous expressions at intermediate stages can thus be derived with the intermediate AUs and the source images. Fig.~\ref{fig:smooth_transistion} shows the continuous editing by Cascade EF-GAN.

\noindent \textbf{Facial Expression Editing on Wild Images:} Editing expression on wild images is much more challenging as the images are captured with complex background and uncontrolled lighting. Our Cascade EF-GAN can be adapted to handle wild images well as illustrated in Fig.~\ref{fig:exp_wild}, where the Cascade EF-GAN is first pre-trained on RaFD and CFEED images and then fine-tuned with wild expressive images from AffectNet~\cite{mollahosseini2017affectnet}. As Fig.~\ref{fig:exp_wild} shows, Cascade EF-GAN can transform the expressions successfully while maintaining the expression-unrelated information unchanged.

\section{Conclusion}
This paper presents a novel {\it Cascade Expression Focal GAN (Cascade EF-GAN)} for realistic facial expression editing. EF-GAN is designed by incorporating three local focuses on eyes, noses and mouths to obtain better preserved identity-related features and details. Such identity-related features reduce model's identity uncertainty, resulting in clearer and sharper facial expression images. In addition, the proposed Cascade EF-GAN performs expression editing in a progressive manner, decomposing large-gap expression changes into multiple small ones. It is therefore more robust in realistic transformation of large-gap facial expressions. Extensive experiments over two publicly available facial expression datasets show that the proposed Cascade EF-GAN achieves superior expression editing as compared with state-of-the-art techniques. 
We expect that Cascade EF-GAN will inspire new insights and attract more interests for better facial expression editing in the near future.


{\small
\bibliographystyle{ieee_fullname}
\bibliography{bibfile}

\begin{thebibliography}{10}\itemsep=-1pt

\bibitem{althoff1999eye}
Robert~R Althoff and Neal~J Cohen.
\newblock Eye-movement-based memory effect: a reprocessing effect in face
  perception.
\newblock {\em Journal of Experimental Psychology: Learning, Memory, and
  Cognition}, 25(4):997, 1999.

\bibitem{baltrusaitis2018openface}
Tadas Baltrusaitis, Amir Zadeh, Yao~Chong Lim, and Louis-Philippe Morency.
\newblock Openface 2.0: Facial behavior analysis toolkit.
\newblock In {\em 2018 13th IEEE International Conference on Automatic Face \&
  Gesture Recognition (FG 2018)}, pages 59--66. IEEE, 2018.

\bibitem{brock2018large}
Andrew Brock, Jeff Donahue, and Karen Simonyan.
\newblock Large scale gan training for high fidelity natural image synthesis.
\newblock {\em arXiv preprint arXiv:1809.11096}, 2018.

\bibitem{chen2019semantic}
Ying-Cong Chen, Xiaohui Shen, Zhe Lin, Xin Lu, I Pao, Jiaya Jia, et~al.
\newblock Semantic component decomposition for face attribute manipulation.
\newblock In {\em Proceedings of the IEEE Conference on Computer Vision and
  Pattern Recognition}, pages 9859--9867, 2019.

\bibitem{chen2019homomorphic}
Ying-Cong Chen, Xiaogang Xu, Zhuotao Tian, and Jiaya Jia.
\newblock Homomorphic latent space interpolation for unpaired image-to-image
  translation.
\newblock In {\em Proceedings of the IEEE Conference on Computer Vision and
  Pattern Recognition}, pages 2408--2416, 2019.

\bibitem{choi2018stargan}
Yunjey Choi, Minje Choi, Munyoung Kim, Jung-Woo Ha, Sunghun Kim, and Jaegul
  Choo.
\newblock Stargan: Unified generative adversarial networks for multi-domain
  image-to-image translation.
\newblock In {\em Proceedings of the IEEE Conference on Computer Vision and
  Pattern Recognition}, pages 8789--8797, 2018.

\bibitem{ding2018exprgan}
Hui Ding, Kumar Sricharan, and Rama Chellappa.
\newblock Exprgan: Facial expression editing with controllable expression
  intensity.
\newblock In {\em Thirty-Second AAAI Conference on Artificial Intelligence},
  2018.

\bibitem{du2014compound}
Shichuan Du, Yong Tao, and Aleix~M Martinez.
\newblock Compound facial expressions of emotion.
\newblock {\em Proceedings of the National Academy of Sciences},
  111(15):E1454--E1462, 2014.

\bibitem{friesen1978facial}
E Friesen and Paul Ekman.
\newblock Facial action coding system: a technique for the measurement of
  facial movement.
\newblock {\em Palo Alto}, 3, 1978.

\bibitem{goodfellow2014generative}
Ian Goodfellow, Jean Pouget-Abadie, Mehdi Mirza, Bing Xu, David Warde-Farley,
  Sherjil Ozair, Aaron Courville, and Yoshua Bengio.
\newblock Generative adversarial nets.
\newblock In {\em Advances in neural information processing systems}, pages
  2672--2680, 2014.

\bibitem{gulrajani2017improved}
Ishaan Gulrajani, Faruk Ahmed, Martin Arjovsky, Vincent Dumoulin, and Aaron~C
  Courville.
\newblock Improved training of wasserstein gans.
\newblock In {\em Advances in Neural Information Processing Systems}, pages
  5767--5777, 2017.

\bibitem{heusel2017gans}
Martin Heusel, Hubert Ramsauer, Thomas Unterthiner, Bernhard Nessler, and Sepp
  Hochreiter.
\newblock Gans trained by a two time-scale update rule converge to a local nash
  equilibrium.
\newblock In {\em Advances in Neural Information Processing Systems}, pages
  6626--6637, 2017.

\bibitem{hsiao2008two}
Janet Hui-wen Hsiao and Garrison Cottrell.
\newblock Two fixations suffice in face recognition.
\newblock {\em Psychological science}, 19(10):998--1006, 2008.

\bibitem{huang2017beyond}
Rui Huang, Shu Zhang, Tianyu Li, and Ran He.
\newblock Beyond face rotation: Global and local perception gan for
  photorealistic and identity preserving frontal view synthesis.
\newblock In {\em Proceedings of the IEEE International Conference on Computer
  Vision}, pages 2439--2448, 2017.

\bibitem{huynh2008scope}
Quan Huynh-Thu and Mohammed Ghanbari.
\newblock Scope of validity of psnr in image/video quality assessment.
\newblock {\em Electronics letters}, 44(13):800--801, 2008.

\bibitem{isola2017image}
Phillip Isola, Jun-Yan Zhu, Tinghui Zhou, and Alexei~A Efros.
\newblock Image-to-image translation with conditional adversarial networks.
\newblock In {\em Proceedings of the IEEE conference on computer vision and
  pattern recognition}, pages 1125--1134, 2017.

\bibitem{karras2017progressive}
Tero Karras, Timo Aila, Samuli Laine, and Jaakko Lehtinen.
\newblock Progressive growing of gans for improved quality, stability, and
  variation.
\newblock {\em arXiv preprint arXiv:1710.10196}, 2017.

\bibitem{kim2017learning}
Taeksoo Kim, Moonsu Cha, Hyunsoo Kim, Jung~Kwon Lee, and Jiwon Kim.
\newblock Learning to discover cross-domain relations with generative
  adversarial networks.
\newblock In {\em Proceedings of the 34th International Conference on Machine
  Learning-Volume 70}, pages 1857--1865. JMLR. org, 2017.

\bibitem{kingma2014adam}
Diederik~P Kingma and Jimmy Ba.
\newblock Adam: A method for stochastic optimization.
\newblock {\em arXiv preprint arXiv:1412.6980}, 2014.

\bibitem{lample2017fader}
Guillaume Lample, Neil Zeghidour, Nicolas Usunier, Antoine Bordes, Ludovic
  Denoyer, et~al.
\newblock Fader networks: Manipulating images by sliding attributes.
\newblock In {\em Advances in Neural Information Processing Systems}, pages
  5967--5976, 2017.

\bibitem{langner2010presentation}
Oliver Langner, Ron Dotsch, Gijsbert Bijlstra, Daniel~HJ Wigboldus, Skyler~T
  Hawk, and AD Van~Knippenberg.
\newblock Presentation and validation of the radboud faces database.
\newblock {\em Cognition and emotion}, 24(8):1377--1388, 2010.

\bibitem{ledig2017photo}
Christian Ledig, Lucas Theis, Ferenc Husz{\'a}r, Jose Caballero, Andrew
  Cunningham, Alejandro Acosta, Andrew Aitken, Alykhan Tejani, Johannes Totz,
  Zehan Wang, et~al.
\newblock Photo-realistic single image super-resolution using a generative
  adversarial network.
\newblock In {\em Proceedings of the IEEE conference on computer vision and
  pattern recognition}, pages 4681--4690, 2017.

\bibitem{li2016deep}
Mu Li, Wangmeng Zuo, and David Zhang.
\newblock Deep identity-aware transfer of facial attributes.
\newblock {\em arXiv preprint arXiv:1610.05586}, 2016.

\bibitem{li2018deep}
Shan Li and Weihong Deng.
\newblock Deep facial expression recognition: A survey.
\newblock {\em arXiv preprint arXiv:1804.08348}, 2018.

\bibitem{li2017generative}
Yijun Li, Sifei Liu, Jimei Yang, and Ming-Hsuan Yang.
\newblock Generative face completion.
\newblock In {\em Proceedings of the IEEE Conference on Computer Vision and
  Pattern Recognition}, pages 3911--3919, 2017.

\bibitem{liang2017generative}
Xiaodan Liang, Hao Zhang, and Eric~P Xing.
\newblock Generative semantic manipulation with contrasting gan.
\newblock {\em arXiv preprint arXiv:1708.00315}, 2017.

\bibitem{lu2018attribute}
Yongyi Lu, Yu-Wing Tai, and Chi-Keung Tang.
\newblock Attribute-guided face generation using conditional cyclegan.
\newblock In {\em Proceedings of the European Conference on Computer Vision
  (ECCV)}, pages 282--297, 2018.

\bibitem{mirza2014conditional}
Mehdi Mirza and Simon Osindero.
\newblock Conditional generative adversarial nets.
\newblock {\em arXiv preprint arXiv:1411.1784}, 2014.

\bibitem{miyato2018spectral}
Takeru Miyato, Toshiki Kataoka, Masanori Koyama, and Yuichi Yoshida.
\newblock Spectral normalization for generative adversarial networks.
\newblock {\em arXiv preprint arXiv:1802.05957}, 2018.

\bibitem{mollahosseini2017affectnet}
Ali Mollahosseini, Behzad Hasani, and Mohammad~H Mahoor.
\newblock Affectnet: A database for facial expression, valence, and arousal
  computing in the wild.
\newblock {\em IEEE Transactions on Affective Computing}, 10(1):18--31, 2017.

\bibitem{pathak2016context}
Deepak Pathak, Philipp Krahenbuhl, Jeff Donahue, Trevor Darrell, and Alexei~A
  Efros.
\newblock Context encoders: Feature learning by inpainting.
\newblock In {\em Proceedings of the IEEE conference on computer vision and
  pattern recognition}, pages 2536--2544, 2016.

\bibitem{pumarola2018ganimation}
Albert Pumarola, Antonio Agudo, Aleix~M Martinez, Alberto Sanfeliu, and
  Francesc Moreno-Noguer.
\newblock Ganimation: Anatomically-aware facial animation from a single image.
\newblock In {\em Proceedings of the European Conference on Computer Vision
  (ECCV)}, pages 818--833, 2018.

\bibitem{qiao2018geometry}
Fengchun Qiao, Naiming Yao, Zirui Jiao, Zhihao Li, Hui Chen, and Hongan Wang.
\newblock Geometry-contrastive gan for facial expression transfer.
\newblock {\em arXiv preprint arXiv:1802.01822}, 2018.

\bibitem{sajjadi2017enhancenet}
Mehdi~SM Sajjadi, Bernhard Scholkopf, and Michael Hirsch.
\newblock Enhancenet: Single image super-resolution through automated texture
  synthesis.
\newblock In {\em Proceedings of the IEEE International Conference on Computer
  Vision}, pages 4491--4500, 2017.

\bibitem{saxe2013exact}
Andrew~M Saxe, James~L McClelland, and Surya Ganguli.
\newblock Exact solutions to the nonlinear dynamics of learning in deep linear
  neural networks.
\newblock {\em arXiv preprint arXiv:1312.6120}, 2013.

\bibitem{shen2017learning}
Wei Shen and Rujie Liu.
\newblock Learning residual images for face attribute manipulation.
\newblock In {\em Proceedings of the IEEE Conference on Computer Vision and
  Pattern Recognition}, pages 4030--4038, 2017.

\bibitem{song2018geometry}
Lingxiao Song, Zhihe Lu, Ran He, Zhenan Sun, and Tieniu Tan.
\newblock Geometry guided adversarial facial expression synthesis.
\newblock In {\em 2018 ACM Multimedia Conference on Multimedia Conference},
  pages 627--635. ACM, 2018.

\bibitem{szegedy2017inception}
Christian Szegedy, Sergey Ioffe, Vincent Vanhoucke, and Alexander~A Alemi.
\newblock Inception-v4, inception-resnet and the impact of residual connections
  on learning.
\newblock In {\em Thirty-First AAAI Conference on Artificial Intelligence},
  2017.

\bibitem{tran2017disentangled}
Luan Tran, Xi Yin, and Xiaoming Liu.
\newblock Disentangled representation learning gan for pose-invariant face
  recognition.
\newblock In {\em Proceedings of the IEEE Conference on Computer Vision and
  Pattern Recognition}, pages 1415--1424, 2017.

\bibitem{wang2020suppressing}
Kai Wang, Xiaojiang Peng, Jianfei Yang, Shijian Lu, and Yu Qiao.
\newblock Suppressing uncertainties for large-scale facial expression
  recognition.
\newblock In {\em CVPR}, 2020.

\bibitem{wang2018high}
Ting-Chun Wang, Ming-Yu Liu, Jun-Yan Zhu, Andrew Tao, Jan Kautz, and Bryan
  Catanzaro.
\newblock High-resolution image synthesis and semantic manipulation with
  conditional gans.
\newblock In {\em Proceedings of the IEEE Conference on Computer Vision and
  Pattern Recognition}, pages 8798--8807, 2018.

\bibitem{wang2018cascaded}
Zheng Wang, Mang Ye, Fan Yang, Xiang Bai, and Shin'ichi Satoh.
\newblock Cascaded sr-gan for scale-adaptive low resolution person
  re-identification.
\newblock In {\em IJCAI}, pages 3891--3897, 2018.

\bibitem{xiao2018elegant}
Taihong Xiao, Jiapeng Hong, and Jinwen Ma.
\newblock Elegant: Exchanging latent encodings with gan for transferring
  multiple face attributes.
\newblock In {\em Proceedings of the European Conference on Computer Vision
  (ECCV)}, pages 168--184, 2018.

\bibitem{yan2018shift}
Zhaoyi Yan, Xiaoming Li, Mu Li, Wangmeng Zuo, and Shiguang Shan.
\newblock Shift-net: Image inpainting via deep feature rearrangement.
\newblock In {\em Proceedings of the European Conference on Computer Vision
  (ECCV)}, pages 1--17, 2018.

\bibitem{yeh2017semantic}
Raymond~A Yeh, Chen Chen, Teck Yian~Lim, Alexander~G Schwing, Mark
  Hasegawa-Johnson, and Minh~N Do.
\newblock Semantic image inpainting with deep generative models.
\newblock In {\em Proceedings of the IEEE Conference on Computer Vision and
  Pattern Recognition}, pages 5485--5493, 2017.

\bibitem{yin2017towards}
Xi Yin, Xiang Yu, Kihyuk Sohn, Xiaoming Liu, and Manmohan Chandraker.
\newblock Towards large-pose face frontalization in the wild.
\newblock In {\em Proceedings of the IEEE International Conference on Computer
  Vision}, pages 3990--3999, 2017.

\bibitem{yu2018generative}
Jiahui Yu, Zhe Lin, Jimei Yang, Xiaohui Shen, Xin Lu, and Thomas~S Huang.
\newblock Generative image inpainting with contextual attention.
\newblock In {\em Proceedings of the IEEE Conference on Computer Vision and
  Pattern Recognition}, pages 5505--5514, 2018.

\bibitem{zeng2018facial}
Jiabei Zeng, Shiguang Shan, and Xilin Chen.
\newblock Facial expression recognition with inconsistently annotated datasets.
\newblock In {\em Proceedings of the European conference on computer vision
  (ECCV)}, pages 222--237, 2018.

\bibitem{zhang2018joint}
Feifei Zhang, Tianzhu Zhang, Qirong Mao, and Changsheng Xu.
\newblock Joint pose and expression modeling for facial expression recognition.
\newblock In {\em Proceedings of the IEEE Conference on Computer Vision and
  Pattern Recognition}, pages 3359--3368, 2018.

\bibitem{zhang2018generative}
Gang Zhang, Meina Kan, Shiguang Shan, and Xilin Chen.
\newblock Generative adversarial network with spatial attention for face
  attribute editing.
\newblock In {\em Proceedings of the European Conference on Computer Vision
  (ECCV)}, pages 417--432, 2018.

\bibitem{zhao20183d}
Jian Zhao, Lin Xiong, Yu Cheng, Yi Cheng, Jianshu Li, Li Zhou, Yan Xu,
  Jayashree Karlekar, Sugiri Pranata, Shengmei Shen, et~al.
\newblock 3d-aided deep pose-invariant face recognition.
\newblock In {\em IJCAI}, volume~2.

\bibitem{zhao2017dual}
Jian Zhao, Lin Xiong, Panasonic~Karlekar Jayashree, Jianshu Li, Fang Zhao,
  Zhecan Wang, Panasonic~Sugiri Pranata, Panasonic~Shengmei Shen, Shuicheng
  Yan, and Jiashi Feng.
\newblock Dual-agent gans for photorealistic and identity preserving profile
  face synthesis.
\newblock In {\em Advances in Neural Information Processing Systems}, pages
  66--76, 2017.

\bibitem{zhao2016peak}
Xiangyun Zhao, Xiaodan Liang, Luoqi Liu, Teng Li, Yugang Han, Nuno Vasconcelos,
  and Shuicheng Yan.
\newblock Peak-piloted deep network for facial expression recognition.
\newblock In {\em European conference on computer vision}, pages 425--442.
  Springer, 2016.

\bibitem{zhu2017unpaired}
Jun-Yan Zhu, Taesung Park, Phillip Isola, and Alexei~A Efros.
\newblock Unpaired image-to-image translation using cycle-consistent
  adversarial networks.
\newblock In {\em Proceedings of the IEEE International Conference on Computer
  Vision}, pages 2223--2232, 2017.

\end{thebibliography}
}

\clearpage

\section{Loss Function}
The loss function for training the proposed EF-GAN contains five terms:
1) the adversarial loss for improving the photo-realism of the synthesized facial expression images to make them indistinguishable from real samples;
2) the conditional expression loss to ensure generated facial expression images to align with the provided target AUs labels;
3) the content loss for preserving the identity information and consistency of the image content.
4) the attention loss to encourage the attentive module to produce sparse attention map and pay attention to the regions that really need modification. 
5) the interpolation loss to constrain the interpolated AUs label has desired sematical meaning and resides on the manifold of natural AUs.

Formally, given a facial expression image and its corresponding local regions $I_x = \{I_{face}, I_{eyes}, I_{nose}, I_{mouth} \}$ with AUs label $y_x$ and the expression residual $r$.
The target expression AUs label $y_z$ is generated by the Interpolator $y_z = Interp(y_x, r)$. The discriminator $D_{interp}$ is trained to distinguish real/fake AUs.
The initial output produced by the Expression Transformer is $I^{init}_{z} = ET(I_x, {y_z})$, where $I^{init}_{z} = \{I^{init}_{face}, I^{init}_{eyes}, I^{init}_{nose}, I^{init}_{mouth} \}$ and $ET = \{ET_{face}, ET_{eye}, ET_{nose}, ET_{mouth} \}$.
We then feed the initial outputs to the refiner, and the final output is generated by $I_{z} = R(I^{init}_{z})$. We define $I = \{I_{z}, I^{init}_{z}\}$ to simplify the expression in the following section.
The discriminator $D$ distinguishes whether the query image is a real image or not. 
To improve the quality of the synthesized image, We introduce a hierarchical $D = \{D_{final}, D_{init} \}$, where $D_{init} = \{D_{face}, D_{eye}, D_{nose}, D_{mouth} \}$ is a set of four discriminators for the initial outputs.
$D_{final}$ examines the final output to judge the holistic of facial features and predict the AUs label, while $D_{init}$ examine the quality of initial outputs.

\noindent {\textbf{Adversarial Loss}}
We adopt the WGAN-GP~\cite{gulrajani2017improved} to learn the parameters.
The adversarial loss function is formulated as:
\begin{equation}
    \begin{aligned}
        \mathcal{L}_{adv} = 
                 & \sum_{i} \{
                    {\mathbb{E}}_{I_{x_i} \sim P_{data}} [{\rm log}D_{i}(I_{x_i})]
                    - {\mathbb{E}}_{I_{i} \sim P_{S}} [{\rm log}D_{i}(I_{i})]  \\
                 & - \lambda_{gp} {\mathbb{E}}_{\tilde{I}_i \sim P_{\tilde{I}_i}} 
                    [\| \nabla_{\tilde{I}_i} D_{i}(\tilde{I}_i) \|_2 - 1]^2
                    \},
        \label{formula:WGAN-GP}
    \end{aligned}
\end{equation}
\noindent where $D_{i} \in D$, $I_{x_i} \in I_x$, $I_i \in I$, $P_{data}$ stands for the data distribution of the real images, $P_{S}$ the distribution of the synthesized images and $P_{\tilde{I}_i}$ the random interpolation distribution. $\lambda_{gp}$ is set to be 10.

\noindent {\textbf{Conditional Expression Loss}}
For a given input $I_{x}$ and the target expression label ${y_z}$, our goal is to synthesize an output image $I_{z}$ with the desired target expression.
To achieve this condition, we add an auxiliary classifier on top of D and impose AUs regression loss when training the network.
In particular, the objective is composed of two terms: an AUs regression loss with generated images used to optimize the parameters of Expression Transformer and Refiner,
and an AUs regression loss of real images used to optimize Discriminator $D_{final}$.
In detail, the loss is formulated as:
\begin{equation}
    \begin{aligned}
        \mathcal{L}_{cond} = 
                   & {\mathbb{E}}_{I_{x} \sim P_{data}} [ \| D_{final}(I_{x}) - y_x \|_2^2] \\
                   & + {\mathbb{E}}_{I_{z} \sim P_{S}} [ \| D_{final}(I_{z}) - y_z \|_2^2].
        \label{formula:condition_loss}
    \end{aligned}
\end{equation}

\noindent {\textbf{Content Loss}}
In order to guarantee that the face in both the input and output images correspond to the same person, we adopt cycle loss~\cite{zhu2017unpaired} to force the model to maintain the identity information and personal content after the expression editing process by minimizing the $L1$ difference between the original image and its reconstruction:
\begin{equation}
    \mathcal{L}_{cont} = {\mathbb{E}}_{I_{x} \sim P_{data}}
                         [ \| I_{rec}  - I_{x}\|_1].
    \label{formula:content_loss}
\end{equation}
Note that the content loss is only applied to original input and final output image.

\noindent {\textbf{Attention Loss}}
To encourage the attentive module to produce sparse attention map and pay attention to the regions that really need modification rather than the whole image, we introduce a sparse loss over the attention map:
\begin{equation}
    \begin{aligned}
        \mathcal{L}_{attn} = {\mathbb{E}}_{x \in X} [\| M_A(I_{face}) \|_2
                             + \| M_A(I_{eye}) \|_2 \\
                             + \| M_A(I_{nose}) \|_2
                             + \| M_A(I_{mouth}) \|_2
                             ].
        \label{formula:attention_loss}
    \end{aligned}
\end{equation}

\noindent {\textbf{Interpolation Loss}}
The interpolation loss is consist of two terms: the regression term to make it has the similar semantic meaning with the pseudo AUs target $y_{p}$ (generated by linear interpolation), and the regularized term to constrain it reside on the manifold of natural AUs:
\begin{equation}
    \begin{aligned}
        \mathcal{L}_{interp} = {\mathbb{E}}_{\hat{y} \sim P_{I}} [\| \hat{y} - y_{p} \|_2
                              +\lambda_{int} {\mathbb{E}}_{\hat{y} \sim P_{R}} [logD_{interp}(\hat{y})],
        \label{formula:interpolation_loss}
    \end{aligned}
\end{equation}

\noindent where $\hat{y}$ stands for the interpolated AUs, $P_{I}$ the data distribution of the interpolated AUs and $P_{R}$ the distribution of the real AUs. $D_{interp}$ is the discriminator for AUs, which is also trained with WGAN-AP. $\lambda_{int}$ is set to be 0.1.

\noindent {\textbf{Overall Objective Function}}
Finally, the overall objective function is expressed as:
\begin{equation}
    \begin{aligned}
        \mathcal{L} = \mathcal{L}_{adv}
                        +\lambda_{1} \mathcal{L}_{cond}
                        +\lambda_{2} \mathcal{L}_{cont}
                        +\lambda_{3} \mathcal{L}_{attn}
                        +\lambda_{4} \mathcal{L}_{interp}
        \label{formula:overall_loss}
    \end{aligned}
\end{equation}
where $\lambda_{1}$, $\lambda_{2}$, $\lambda_{3}$ and $\lambda_{4}$ are the hyper-parameters that control the relative importance of every loss term. 

In Cascade EF-GAN, the total loss is the sum of the loss of each EF-GAN with equal weights.

\section{Network Architecture}
\label{sec:network_arc}
\begin{figure*}[t]
\begin{center}
\includegraphics[width=1\linewidth, height=650pt]{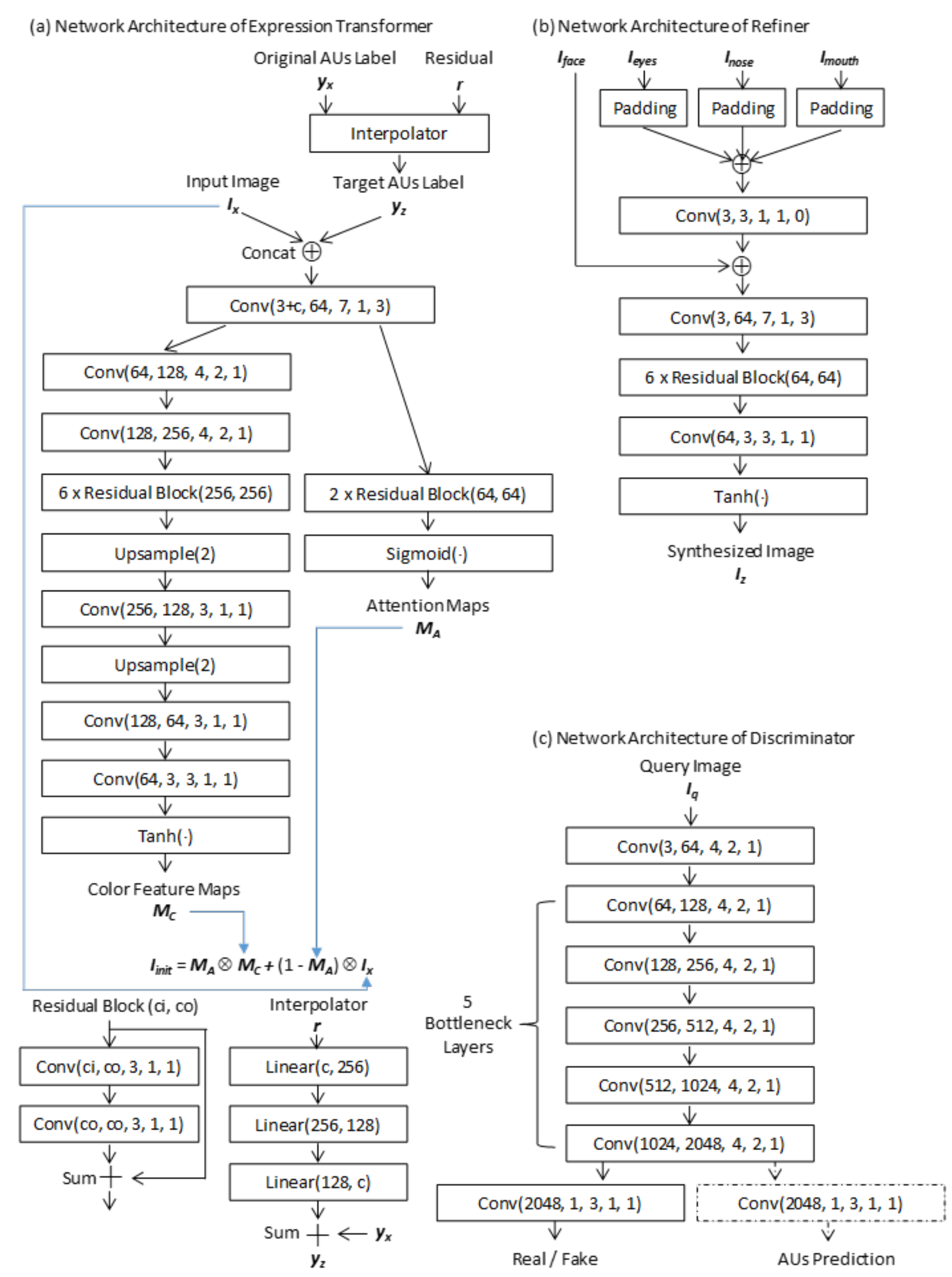}
\end{center}
\caption{
    The detailed network architecture of our proposed model. (a-c) shows the architecture of Expression Transformer, Refiner and Discriminator, respectively. Conv(Nin, Nout, k, s, p) denotes a convolutional layer whose input channel number is Nin, output channel number is Nout, kernel size is k, stride is y and padding is p. Linear(Nin, Nout) denotes a fully connected layer with Nin and Nout as its input and output channel number, respectively. Parameter c denotes dimension of AUs label.
}
\label{fig:detailed_arc}
\end{figure*}

In EF-GAN, We have one global branch that captures the global facial structures and three local focuses branches that help better preserve identity-related features as well as local details around eyes, noses and mouths. Each branch shares similar architecture without sharing weights.
The detailed network architecture is illustrated in Fig.~\ref{fig:detailed_arc}. 
Note that all the convolutional layers in Expression Transformer and Refiner are followed by Instance Normalization layer and ReLU activation layer (omitted in the figure for simplicity), except for the output layer.
And the convolutional layers of Discriminator are followed by Leakly ReLU activation layer with slope of 0.01.
The number of bottleneck layers of the discriminator for global face images is set to 5 while that of local regions is set to 3 according to the size of the images.
The AUs prediction layer is only applied to the final output.

We stack multiple EF-GAN modules in sequence to form Cascade EF-GAN.

\section{Training Details}
\noindent {\bf EF-GAN Training Details.}
We adopt Adam optimizer~\cite{kingma2014adam} with ${\beta}_1$ = 0.5, ${\beta}_2$ = 0.999 for EF-GAN optimization. 
We set $\lambda_1, \lambda_2$ $\lambda_3$, $\lambda_4$ to be 3000, 10, 0.1 and 1 to balance the magnitude of the losses.
The batchsize is set to 2. The total number of epochs is set to 100. 
The initial learning rate is set to 1e-4 for the first 50 epochs, and linearly decay to 0 over the remaining epochs.
Beyond that, we apply Orthogonal Initialization~\cite{saxe2013exact} and Spectral Normalization~\cite{miyato2018spectral} in all convolutional layers except the output layer to stabilize the training process.

\medskip
\noindent {\bf Cascade EF-GAN Training Details.}
To train the Cascade EF-GAN, we first use the weights of a well-trained EF-GAN model to initialize each EF-GAN module in the cascade. Then we train the Cascade EF-GAN model end-to-end for the first 10 epochs, with learning rate starting from 1e-5 and linearly decayed to 0 over the remaining epochs.

\medskip
\noindent{\bf Training Time.}
We use a Tesla V100 GPU in training. For RaFD dataset, it takes 13 hours in training EF-GAN and 8 hours for fine-tuning the Cascade EF-GAN structure. For CFEED dataset, it takes 33 and 20 hours in training and fine-tuning, respectively.

\section{More Results}
We have also presented more results generated by our proposed Cascade EF-GAN in the following pages.

\begin{figure*}[t]
\begin{center}
\includegraphics[width=1\linewidth]{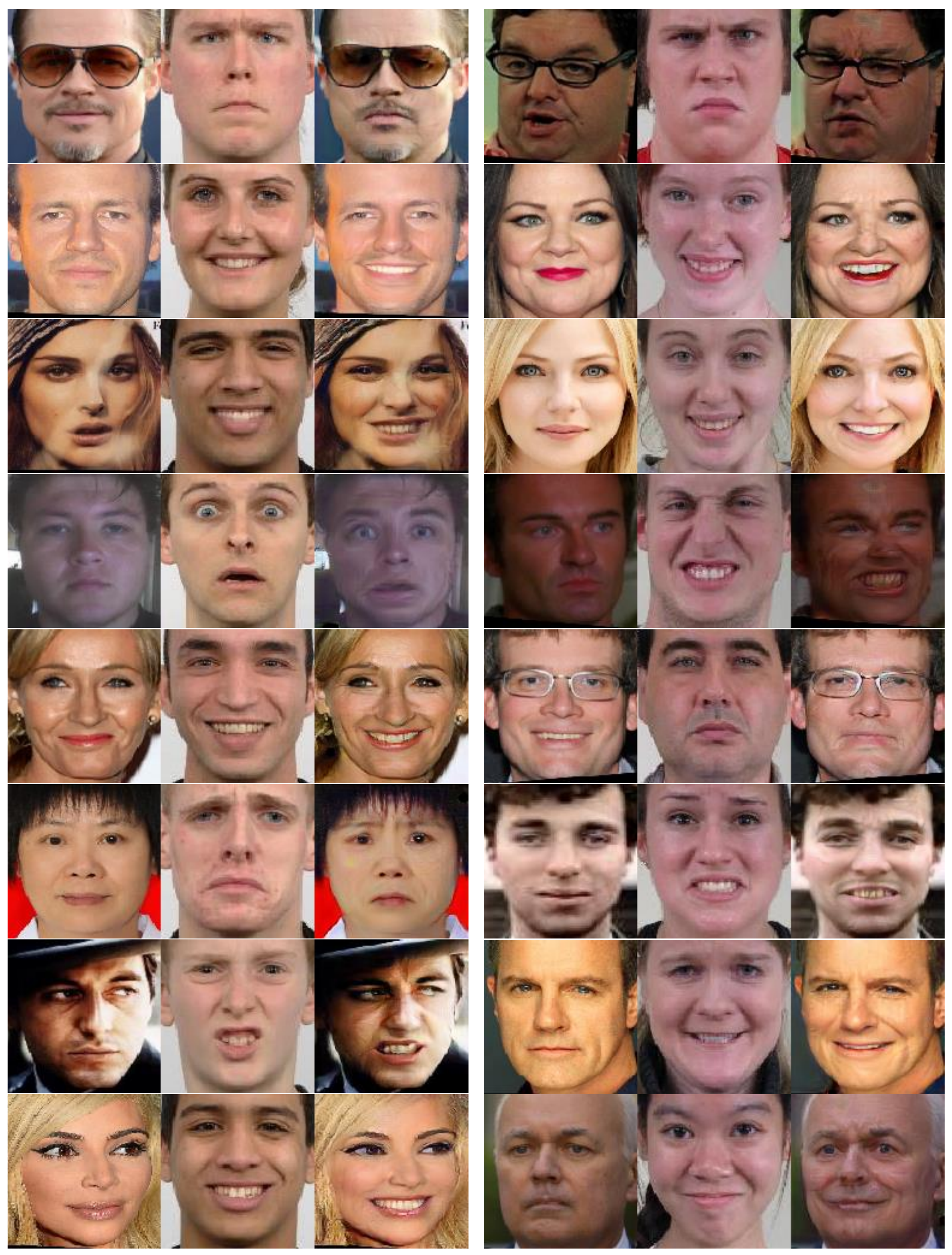}
\end{center}
\caption{
    Additional expression editing results on wild images. In each triplet, the first column is input facial image, the second column is the image with desired expression and the last column is the synthesized result.
}
\end{figure*}

\begin{figure*}[t]
\begin{center}
\includegraphics[width=1\linewidth]{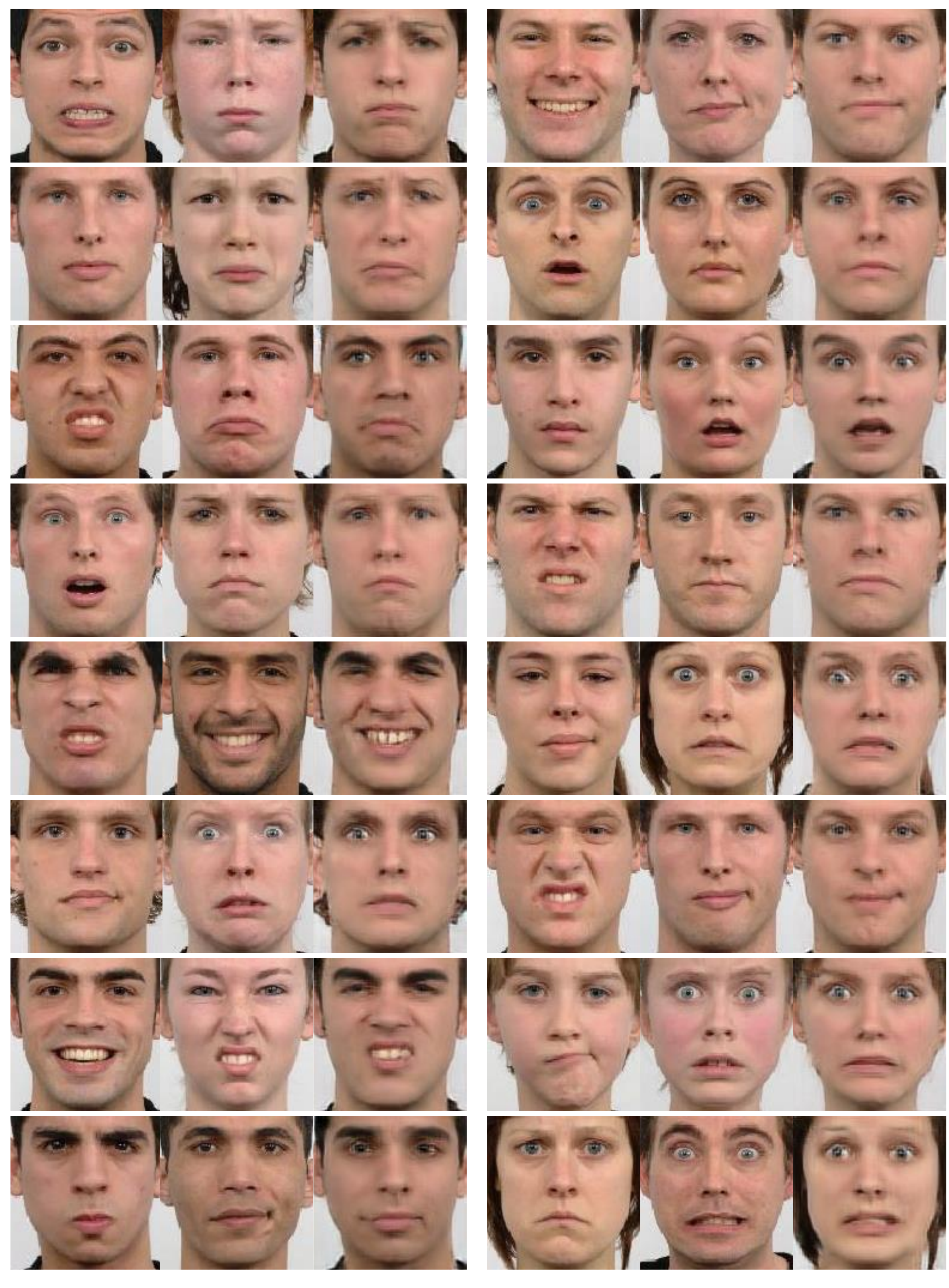}
\end{center}
\caption{
    Additional expression editing results on RaFD. In each triplet, the first column is input facial image, the second column is the image with desired expression and the last column is the synthesized result.
}
\end{figure*}


\begin{figure*}[t]
\begin{center}
\includegraphics[width=1\linewidth]{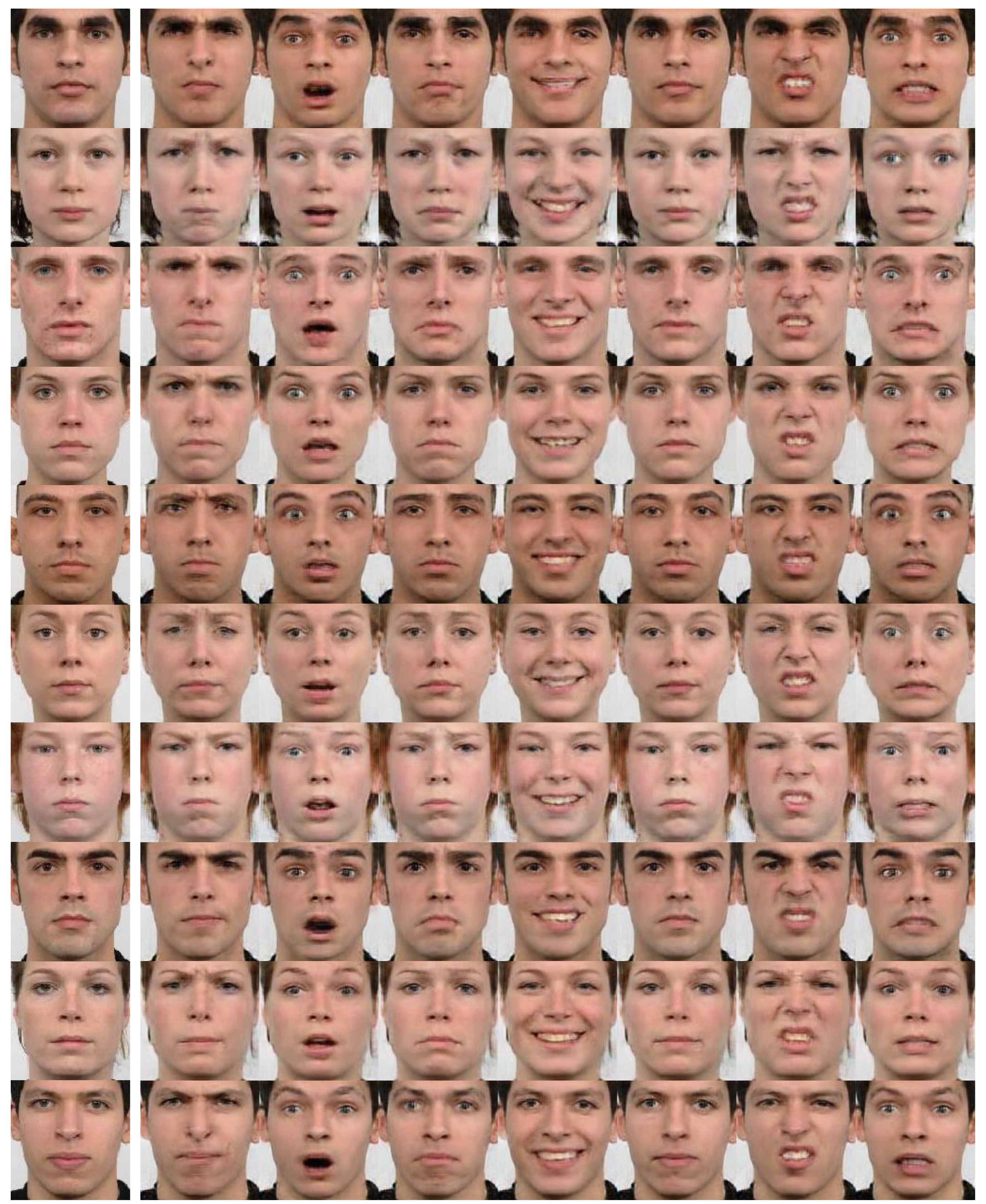}
\end{center}
\caption{
    Additional expression editing results on RaFD (Input, Angry, Surprised, Sad, Happy, Neutral, Disgusted, Fearful).
}
\end{figure*}

\begin{figure*}[t]
\begin{center}
\includegraphics[width=1\linewidth]{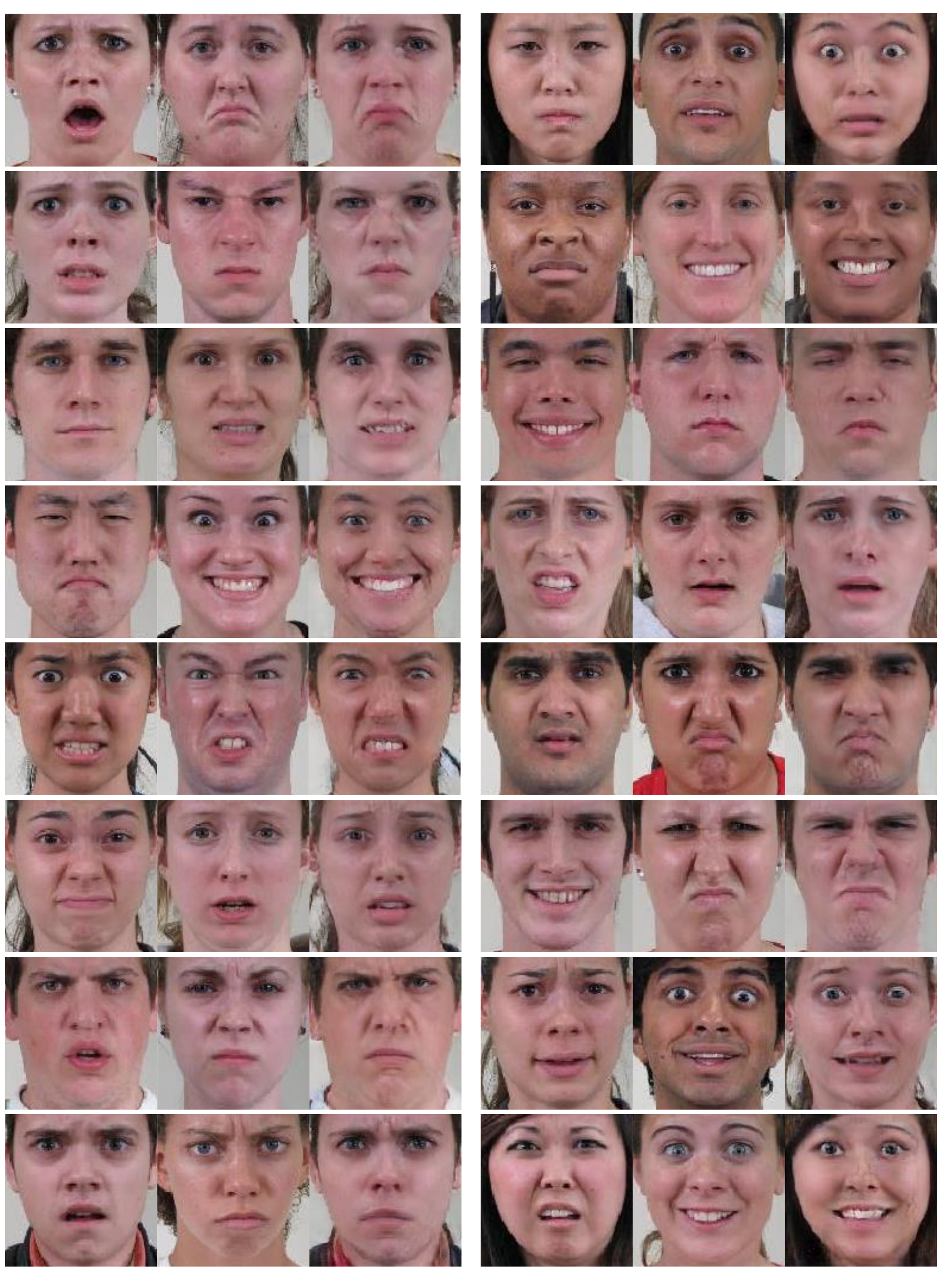}
\end{center}
\caption{
    Additional expression editing results on CFEED. In each triplet, the first column is input facial image, the second column is the image with desired expression and the last column is the synthesized result.
}
\end{figure*}


\begin{figure*}[t]
\begin{center}
\includegraphics[width=1\linewidth]{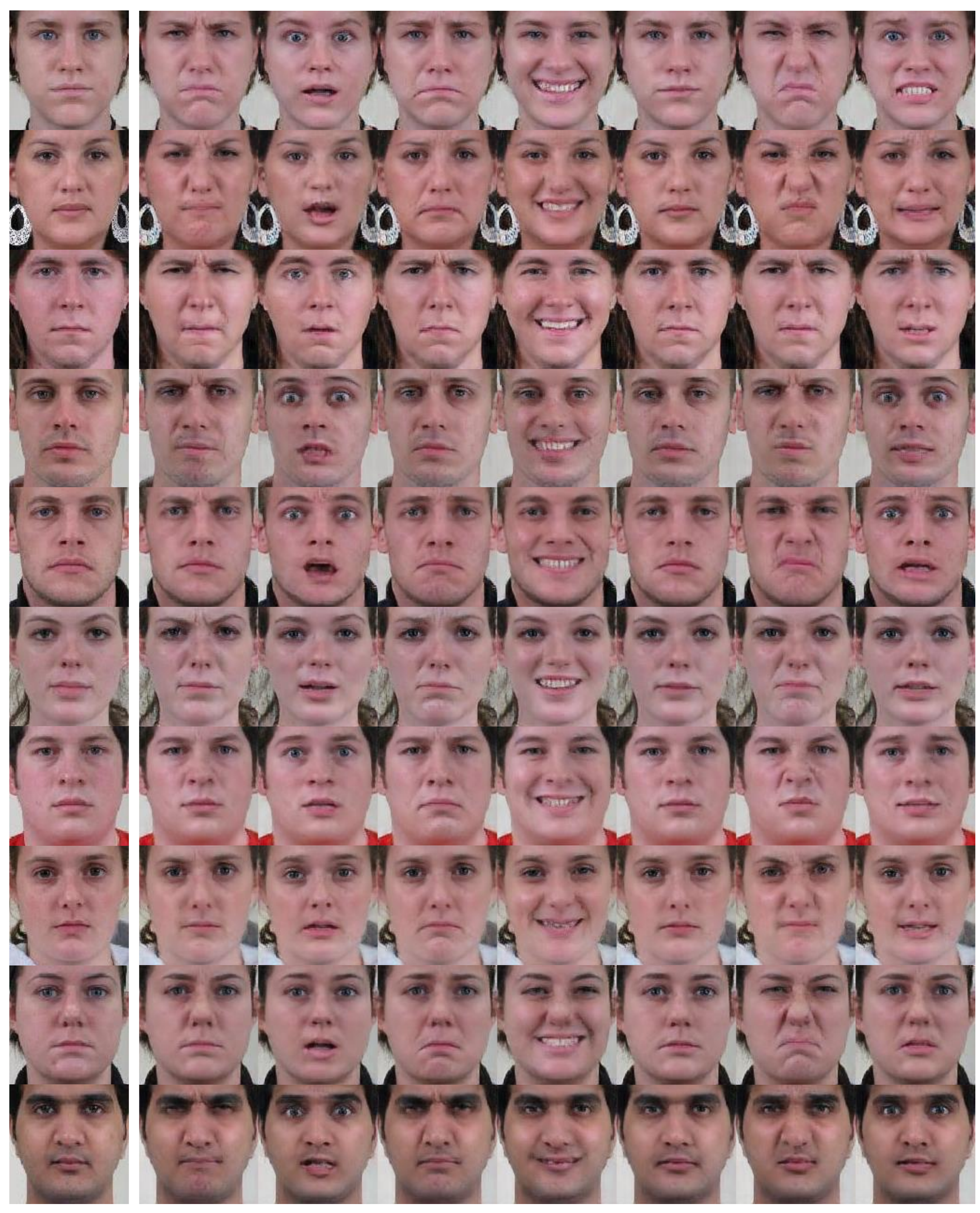}
\end{center}
\caption{
    Additional expression editing results on CFEED (Input, Angry, Surprised, Sad, Happy, Neutral, Disgusted, Fearful).
}
\end{figure*}

\end{document}